\pdfoutput=1
\documentclass[11pt]{article}

\usepackage[final]{acl}

\usepackage{times}
\usepackage{latexsym}

\usepackage[T1]{fontenc}
\usepackage[utf8]{inputenc}

\usepackage{microtype}

\usepackage{inconsolata}

\usepackage{graphicx}
\usepackage{xcolor}         % colors
\usepackage{amsmath}
\usepackage{adjustbox}
\usepackage{graphicx}
\usepackage{subcaption}
\usepackage{booktabs}
\usepackage{multirow}
\usepackage{amsmath}
\usepackage{float}
\usepackage{pifont}
\usepackage{amssymb}
\usepackage{bbding}
\usepackage{latexsym}
\usepackage{wrapfig}
\usepackage{threeparttable}
\usepackage{tcolorbox}
\usepackage{booktabs,amsfonts,dcolumn}
\usepackage{tabularx}
\usepackage{colortbl}

\newcommand{\headercolor}{\rowcolor{gray!15}}
\newcommand{\ours}{PA-RAG}

\title{PA-RAG: RAG Alignment via Multi-Perspective Preference Optimization}
\author{
Jiayi Wu\textsuperscript{1},
Hengyi Cai\textsuperscript{2}, 
Lingyong Yan\textsuperscript{3},
Hao Sun\textsuperscript{4},  \\
\textbf{ 
Xiang Li\textsuperscript{1}\thanks{Corresponding Author},
Shuaiqiang Wang\textsuperscript{3}, 
Dawei Yin\textsuperscript{3},
Ming Gao\textsuperscript{1}}
\\
\textsuperscript{1}School of Data Science and Engineering, East China Normal University\\
\textsuperscript{2}Chinese Academy of Sciences
\textsuperscript{3}Baidu Inc
\textsuperscript{4}Peking University \\
\tt{jiayiwu@stu.ecnu.edu.cn} \\
}

\begin{document}
\maketitle
\begin{abstract}

The emergence of Retrieval-augmented generation (RAG) has alleviated the issues of outdated and hallucinatory content in the generation of large language models (LLMs), yet it still reveals numerous limitations.
When a general-purpose LLM serves as the RAG generator, it often suffers from inadequate response informativeness, response robustness, and citation quality. 
Past approaches to tackle these limitations, either by incorporating additional steps beyond generating responses or optimizing the generator through supervised fine-tuning (SFT), still failed to align with the RAG requirement thoroughly.
Consequently, optimizing the RAG generator from multiple preference perspectives while maintaining its end-to-end LLM form remains a challenge. To bridge this gap, we propose Multiple Perspective \textbf{P}reference \textbf{A}lignment for \textbf{R}etrieval-\textbf{A}ugmented \textbf{G}eneration (\ours{}), a method for optimizing the RAG generator to align with RAG requirements comprehensively.
Specifically, we construct high-quality instruction fine-tuning data and multi-perspective preference data by sampling varied quality responses from the generator across different prompt documents quality scenarios. Subsequently, we optimize the generator using SFT and Direct Preference Optimization (DPO). Extensive experiments conducted on four question-answer datasets across three LLMs demonstrate that \ours{} can significantly enhance the performance of RAG generators. Our code and datasets are available at \url{https://github.com/wujwyi/PA-RAG}.
\end{abstract}
\section{Introduction} 
\begin{figure}[t]
    \centering
    \includegraphics[width=0.48\textwidth]{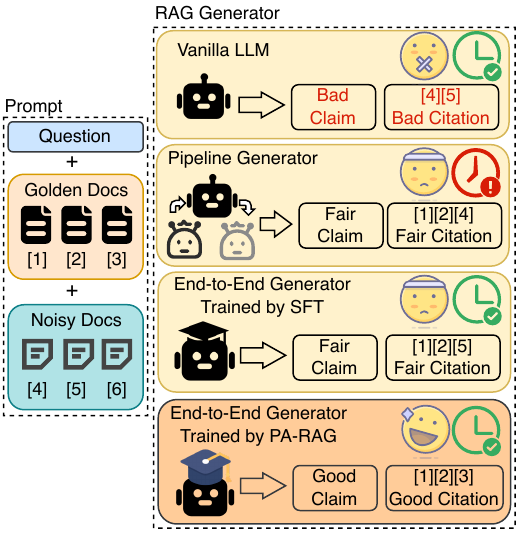}
    \caption{\ours{} retains the end-to-end form of the generator while enabling the generator to learn multi-perspective preference information.}
    \label{fig:intro}
\vspace{-15pt}
\end{figure}

Pre-trained large language models (LLMs) demonstrate impressive question-answering capabilities but also reveal certain drawbacks, such as generating outdated or hallucinatory information \citep{Hallucination_survey,turpin2024language}. The emergence of retrieval-augmented generation (RAG) \citep{gao2023retrieval} has alleviated this issue by incorporating retrieved documents relevant to the question into the prompt, thereby providing the LLMs with information they may not know.
% which enhances the factual accuracy and relevance of the generated responses.

However, while RAG systems help improve factual grounding, the role of the generator in previous work is usually far from well-aligned. 
Despite their impressive pre-training, off-the-shelf LLMs are not optimally suited for the specific requirements of RAG tasks without proper task alignment.
1) \textbf{Helpfulness for RAG application (response informativeness)} \citep{dong2024unsupervised}: The generator should possess a refined ability to identify and utilize valuable information from the provided context. When presented with a mix of high-quality and less relevant information, it should actively and effectively leverage the valuable content while disregarding the rest.
% This ability is particularly important in RAG scenarios where retrieved documents may vary in relevance and quality.
2) \textbf{Harmlessness for RAG application (response robustness)} \citep{yoran2023making,zhu2024atm}: 
% tain noisy information. 
A well-aligned generator should demonstrate robust noise resistance, capable of filtering out misleading information without being adversely affected. This ability ensures the generated responses remain accurate and coherent even when the input context is imperfect.
3) \textbf{Honesty for RAG application (citation quality)} \citep{gao2023enabling,sun2023towards}: The generator should be capable of producing responses that are firmly rooted in the provided documents while maintaining appropriate attribution. This involves explicitly citing the source documents when necessary, and ensuring that the generated content is not only factually accurate but also traceable to the retrieved documents.

% However, several challenges arise if a general LLM that has yet to undergo further optimization is used directly as the generator in an RAG system.
% However, directly employing an off-the-shelf LLM as the generator in a RAG system introduces several challenges.
% First, there is the issue of inadequate response informativeness \citep{dong2024unsupervised}. The generator's response should contain a complete answer. However, the generator may not fully leverage the provided documents, potentially overlooking valuable documents that contain the answer, leading to information gaps in the responses. 
% Second, there is the problem of inadequate response robustness \citep{yoran2023making,zhu2024atm}. The generator needs to be able to resist the interference of noisy documents. The retriever in the RAG system is imperfect and may retrieve documents that are irrelevant to the query or related to the query but do not contain the answer. These noisy documents may distract the generator, resulting in incomplete or incorrect answers. 
% Third, the quality of citations is often inadequate \citep{gao2023enabling,sun2023towards}. The correctness of the generator's response should be verifiable. However, ensuring that the generator not only outputs the correct answers but also accurately cites the corresponding documents remains challenging.

To satisfy the aforementioned RAG requirements, existing RAG systems mainly fall into two categories: end-to-end architecture and pipeline architecture. 
These approaches attempt to bridge the gap between off-the-shelf LLMs and the specific demands of RAG, but both face significant limitations.
% The end-to-end architecture generators are mainly trained by further supervised fine-tuning (SFT) on general LLMs \citep{yoran2023making,fang2024enhancing}. 
End-to-end architectures primarily rely on supervised fine-tuning (SFT) of general LLMs \citep{yoran2023making,fang2024enhancing}. 
These approaches focus on constructing high-quality responses in RAG scenarios. 
% While this approach has demonstrated some success, it struggles with the inherent complexity of RAG tasks. The quality of the retrieved documents can vary dramatically, making it essential for the generator to prioritize useful information and disregard irrelevant or misleading content.
% However, SFT methods typically do not incorporate preference information into the training process, which limits their ability to adapt to this variability \citep{DPO}. 
% As a result, the generator often fails to fully align with RAG-specific requirements, particularly when dealing with noisy or conflicting retrievals.
% However, through further analysis, we found that RAG requirements are pretty complex, and the gap in RAG objectives is significant when the quality of retrieval documents varies. 
However, through further analysis, we found that the requirements for the generators in RAG tasks are highly context-dependent and often interwoven.
% For example, when the retrieved documents are highly relevant and informative, the generator should be able to fully utilize the content.
These varying and intertwined demands make it challenging for a model to meet all RAG objectives through standard SFT alone, as it does not incorporate the necessary preference information required for adapting to different retrieval scenarios.
 % Thus, modeling these complex requirements requires incorporating preference information into the training data.
% Nevertheless, the SFT training method cannot effectively model preference information \citep{DPO}, resulting in the generator not fully aligning with RAG requirements. 
% Pipeline architecture generators involve additional steps beyond generating responses, such as re-rankers for document filtering or verifiers to validate whether citations support claims \citep{dong2024understand,yu2024rankrag,wang2024rear,sun2023towards}.
On the other hand, pipeline architectures introduce additional steps beyond generation, such as re-ranking retrieved documents, filtering irrelevant information, or employing post-hoc verification to ensure that citations support the claims \citep{dong2024understand,yu2024rankrag,wang2024rear,sun2023towards}. 
However, these additional steps can only satisfy certain RAG requirements and still have a considerable gap in aligning with the global RAG requirements.
Moreover, these additional steps introduce extra computational costs and time consumption, making it less practical for large-scale deployment. 
This further inspires us to think:
% \textit{Is it possible to fully align the generator with the RAG requirements while retaining the end-to-end architecture?}
\textit{Is it possible to fully align the generator with the diverse RAG requirements while retaining the simplicity and efficiency of an end-to-end architecture?}

To this end, in this paper, we propose Multi-Perspective \textbf{P}reference \textbf{A}lignment for \textbf{R}etrieval-\textbf{A}ugmented \textbf{G}eneration (\ours{}), aiming to optimize the generator of RAG systems to align comprehensively with specific RAG requirements.
% such as handling noisy retrievals, identifying and leveraging high-quality information, and grounding responses in retrieved documents.
As illustrated in Figure~\ref{fig:intro}, 
% \ours{} retains the end-to-end form of the generator while enabling the generator to learn multi-perspective preference information, thereby fully exploiting the model's potential. 
\ours{} maintains the end-to-end architecture of the generator while enabling it to learn multi-perspective preference information.
The training of \ours{} is divided into two phases.
The first phase is foundational capability training, where the generator acquires the basic ability to utilize and cite documents through instruction fine-tuning. 
To construct high-quality instruction fine-tuning data, we utilize ChatGPT to generate complete and correct answers and employ a \textit{citation rewrite} mechanism to ensure citation quality. 
The second phase is the multi-perspective preference optimization phase, in which the generator is trained using Direct Preference Optimization (DPO)~\citep{DPO} to learn preference information from different perspectives. 
This phase encompasses three sub-stages, sequentially enhancing the generator's response informativeness, response robustness, and citation quality. 
% 什么叫做 "prompt document qualities"
To construct high-quality preference data, we sample responses of varying quality from the generator across scenarios with different document qualities in prompt to build preference data for informativeness and robustness. 
% 这句话和之前提到的 \textit{citation rewrite} 有点重复
Additionally, we employ the \textit{citation rewrite} mechanism to construct preference data for citation quality.

We conducted extensive experiments on four QA datasets and three LLMs, demonstrating that PA-RAG significantly enhances the generator's performance. The improvement achieved by \ours{} far surpasses the performance gains using only SFT or additional steps. LLMs trained with \ours{} show an average absolute improvement of 13.97\% in correctness, 49.77\% in citation recall, and 39.58\% in citation precision. Our contributions can be summarized as follows:
\vspace{-3pt}
\begin{itemize}
    \item We propose \ours{}, which achieves comprehensive alignment of LLMs with specific RAG requirements through a multi-stage, multi-perspective training while preserving its end-to-end architecture.
    \vspace{-1pt}
    \item We publicly release our training data, which includes 58.9k instruction fine-tuning data and 48.7k preference optimization data.
    \vspace{-1pt}
    \item Through extensive experiments, we demonstrate the effectiveness of \ours{}, showing that the preference optimization from each perspective is effective.

\end{itemize}

\section{Preliminaries}
In this section, we describe the optimization objectives of the RAG generator, the motivation for multi-perspective optimization, and the rationale for choosing preference optimization over mere instruction fine-tuning.

\subsection{Optimization Objectives of the RAG Generator}
\label{sec:Optimization Objectives}
As illustrated in Figure~\ref{fig:intro}, when answering a question, the RAG generator outputs the corresponding response after receiving the question and multiple relevant documents retrieved by the retriever. Our optimization goal is to enable the generator to fully utilize valuable documents and accurately cite the references corresponding to claims.

The ability to fully utilize valuable documents corresponds to the correctness of the response, meaning that all answers contained within the documents should be included in the generator’s output. Formally, let $G$ represent the generator, $x$ denotes the input, $D=\{d_1,d_2,...,d_n\}$ represent the documents retrieved by the retriever, $A=\{a_1,a_2,...,a_n\}$ represent the short answers contained in the documents, and $y$ denote the response generated by the generator. The correct response by the generator can be expressed as:
\begin{equation}
\begin{aligned}
    y &= G(x, D) \\
    \text{s.t.} & \forall a_i \in A, C(y, a_i) = \text{True}
\end{aligned}
\end{equation}
where $C(y, a_i) = \text{True}$ implies that the answer $a_i$ is included in $y$.

Based on the generator's ability to correctly answer questions, it is necessary to enhance its capability to accurately cite documents corresponding to claims, referred to as citation quality. Each response $y$ contains multiple statements comprising a claim and multiple citations. 
% We aim for each claim to be fully supported by the documents corresponding to the citations and avoid citing documents unrelated to the claim. 
We aim for each claim to be fully supported by the cited documents, while avoiding irrelevant citations.
Formally, we denote the statement as $s$, the claim as $c$, and the citation as $t$. The correct citation of references corresponding to claims can be expressed as follows:
\begin{equation}
y=\{s_1,s_2,...,s_n\}
\end{equation}
\begin{equation}
s_i=\{\text{``claim'':}c_i, \text{``citation'':}t_i=[t_{i1},t_{i2},...,t_{in}]\}
\end{equation}
\begin{equation}
\forall s_i \in y, \phi(\text{concat}(t_i),c_i)=1
\label{func:4}
\end{equation}

\begin{equation}
\begin{aligned}
& \forall s_i \in y, \forall t_{ij} \in t_i, \\
& \phi(t_{ij}, c_i) = 1 \lor \phi(\text{concat}(t_i \setminus t_{ij}), c_i) = 0
\label{func:5}
\end{aligned}
\end{equation}

Here, $\text{concat}(t_i)$ denotes the concatenation of all cited documents, and $\phi(t,c)=1$ indicates that the claim $c$ is fully supported by the citation $t$.
Formula~\ref{func:4} indicates that all claims can be fully supported by their citations, 
% while Formula~\ref{func:5} illustrates that no documents irrelevant to the claim are cited.
while Formula~\ref{func:5} ensures that no irrelevant documents are cited.

\subsection{Preference Optimization Under Different Document Qualities}
\label{sec 2.2}
As described in \S~\ref{sec:Optimization Objectives}, enhancing the correctness of the RAG generator is the primary objective. 
However, optimizing correctness is highly complex. 
Specifically, the probability of a model correctly answering a question is $P(y|x)$. 
In the RAG framework, 
% the probability of providing a correct answer is related to the input documents, expressed as $P(y|x)=\sum_D[P(D|x) \cdot P(y|D,x)]$. 
this probability depends on the distribution of the retrieved documents and can be formulated as $P(y|x)=\sum_D[P(D|x) \cdot P(y|D,x)]$. 
% Nevertheless, the RAG retriever in a RAG system is not perfect, resulting in variable document quality under different circumstances, leading to a non-uniform distribution of $P(D|x)$. 
% Therefore, it is necessary to optimize the generator under varying document quality conditions.
However, the retriever in a RAG system is not perfect, leading to variability in document quality under different circumstances, which results in a non-uniform distribution of $P(D|x)$. 
Therefore, it is essential to optimize the generator under different document quality conditions.

\begin{figure*}[t]
    \centering
    \includegraphics[width=0.9\textwidth]{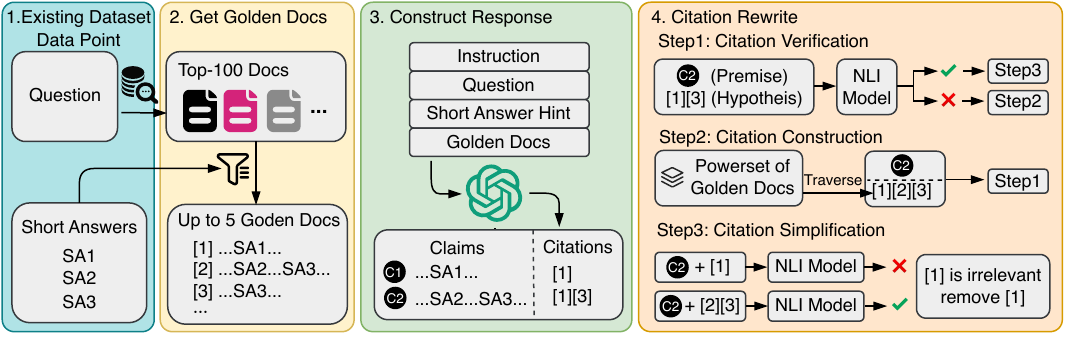}
    \caption{Overview of instruction fine-tuning data construction.}
    \label{fig:SFT data overview}
\vspace{-10pt}
\end{figure*}

We categorize document quality into two types. 
% The first type is where all documents contain answers. 
The first type involves scenarios where all documents contain relevant answers.
In this scenario, we expect the generator to fully utilize the valuable documents and produce complete answers. 
Therefore, when the documents are of high-quality, we need to focus on optimizing response informativeness. 
The second type includes scenarios where the document set contains noisy documents. 
In this case, we expect the generator to avoid the interference of noisy documents and still maintain the completeness of its answers. 
% Thus, when the documents are of low-quality, the emphasis is on optimizing response robustness. 
Thus, when dealing with low-quality documents, the emphasis shifts to optimizing response robustness.
% However, the directions of response informativeness and response robustness are not aligned; 
However, optimizing for informativeness and robustness pulls the generator in different directions.
The former encourages referencing more documents, while the latter demands ignoring more.
% the former focuses on referencing more documents, while the latter emphasizes ignoring more documents. 
% Directly using SFT to optimize the model cannot teach the model to focus on valuable documents and ignore noisy ones (see \S~\ref{sec: dpo vs sft} for details), making it challenging to optimize the generator effectively. 
Standard instruction fine-tuning is insufficient to teach the model how to balance these competing demands (documents encouraged to be referenced are valuable, while those that should be ignored are considered noisy, see \S~\ref{sec: dpo vs sft} for details), making it difficult to optimize the generator effectively.
% Therefore, it is necessary to introduce preference information and perform preference optimization from multiple perspectives on the generator.
To address this challenge, we introduce preference information and perform multi-perspective preference optimization to guide the generator in focusing on valuable documents while ignoring noisy ones.
\section{Methodology}

After defining the optimization objectives, we trained the generator in two phases. The first phase enabled a general LLM to acquire basic RAG capabilities. The method for constructing the training data for this phase is detailed in \S~\ref{sec:IFT}. The second phase involved multi-perspective preference optimization, further enhancing the generator's response informativeness, response robustness, and citation quality. The method for constructing the training data for this phase is described in \S~\ref{sec:DPO}.

\subsection{Instruction Fine-tuning for Basic RAG Capabilities}
\label{sec:IFT}

During the instruction fine-tuning phase, we aimed to equip the generator with fundamental abilities to utilize and cite documents following the optimization objectives outlined in \S~\ref{sec:Optimization Objectives}. When constructing the training data, we employ ChatGPT-3.5 (\texttt{GPT-3.5-Turbo-1106}) and introduce a \textit{citation rewrite} mechanism to create near-perfect responses. An overview of the instruction fine-tuning data construction is illustrated in Figure~\ref{fig:SFT data overview}.

Firstly, we need to acquire high-quality documents after obtaining the questions and short answers provided by existing datasets. We use the RAG retriever to retrieve the top 100 most relevant documents from the retrieval corpus, and then filter to retain the golden documents that contain the short answer. Subsequently, we select up to five documents from all the golden documents as prompt documents, ensuring that all short answers are included in the prompt documents.

Secondly, we need to construct responses that include all short answers and exhibit high citation quality. Specifically, we use ChatGPT-3.5 to generate responses. To enhance the quality of the responses, we include instructions and short answer hints in the ChatGPT prompt. Detailed prompt can be found in Appendix~\ref{appendix:prompt}.
% shorted as follows:

% \begin{tcolorbox}[title=Prompt for ChatGPT-3.5]
% \small
% \textbf{Instruction:} Write an accurate, engaging, and concise answer for the given question ... \\
% \textbf{Qustion:} \{Question\}\\ 
% The final answer should contain the following short answers: \{Short answers\} \\
% \textbf{Documents:} \{Documents\} \\
% \textbf{Answer:}
% \end{tcolorbox}

Although ChatGPT-3.5 is quite powerful and the prompt includes short answer hints, the quality of the citations in the answers is unsatisfactory. Therefore, we introduce a \textit{citation rewrite} mechanism to improve the quality of substandard citations, which is divided into the following steps: \textbf{First}, verify the citation. We use a Natural Language Inference (NLI) model to validate whether the document corresponding to each citation (premise) in the statement can satisfy the claim (hypothesis). \textbf{Second}, construct the citation. If the citation in the first step cannot support the claim, we traverse the power set of all prompt documents as citations to explore a feasible citation scheme. \textbf{Third}, simplify the citation. If the citation can support the claim, we check one by one whether the citation is irrelevant to the claim and remove irrelevant citations.

After generating answers with ChatGPT-3.5 and performing \textit{citation rewrite}, we filter to retain responses that include all short answers and contain only accurate citations to serve as the training data for the instruction fine-tuning of the generator, which is then subsequently fine-tuned.
\begin{figure*}[t]
    \centering
    \includegraphics[width=0.70\textwidth]{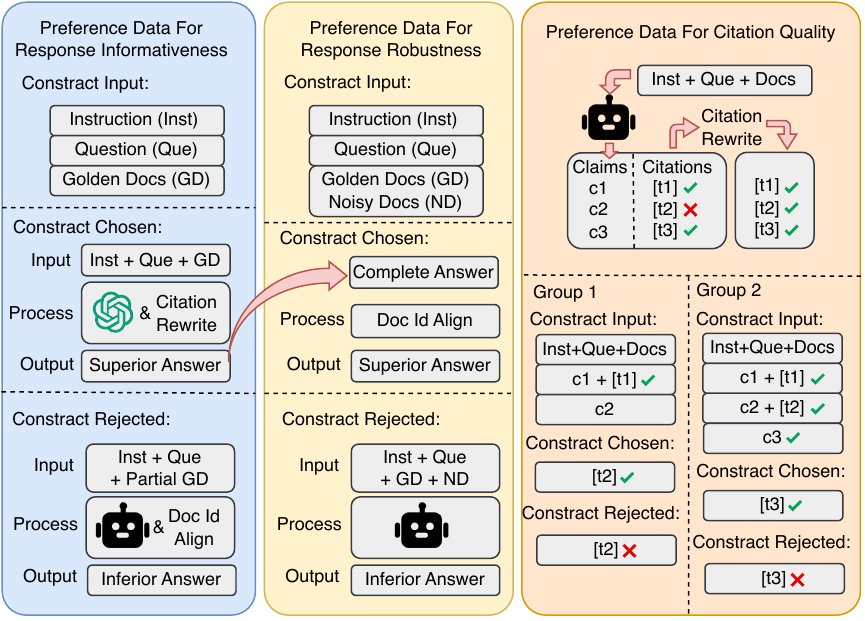}
    \caption{Overview of preference data construction.}
    \label{fig:reference data overview}
\end{figure*}

\subsection{Preference Optimization of the Generator through DPO}
\label{sec:DPO}

Once the generator has acquired the fundamental ability to utilize and reference documents, we sequentially optimize the generator from the perspectives of response informativeness, response robustness, and citation quality during the preference optimization phase. Preference optimization requires constructing data with preference information, including an input, a superior answer (chosen output), and an inferior answer (rejected output). We will progressively explain the methods for constructing preference data from each perspective. An overview of the preference data construction is illustrated in Figure~\ref{fig:reference data overview}.

\subsubsection{Response Informativeness}
\label{sec:RI}

Response informativeness refers to the completeness of the answer in the response. The optimization objective from this perspective is to make the generator fully utilize the document that contains the short answer. 
% As described in section 2.2, we enhance response informativeness in the case of high-quality documents. When constructing preference data, the chosen output is the response that includes all short answers by fully utilizing the golden document, while the rejected output is the response that some answers are lacking or erroneous because the generator has overlooked parts of the golden document.

\textbf{Input Construction:} Similar to constructing the input component of the instruction fine-tuning data, the input includes the instruction, the question, and the high-quality prompt documents containing all short answers from up to 5 golden documents.

\textbf{Chosen Output Construction:} The construction method is consistent with the output component of the instruction fine-tuning data, using ChatGPT-3.5 and the \textit{citation rewrite} mechanism to construct a response that includes all short answers and the accurate citation.

\textbf{Rejected Output Construction:} To simulate the scenario where the generator ignores parts of the golden documents, we delete some golden documents from the prompt, generate responses with the generator, and filter to retain inferior answers that are incomplete. Since some documents are removed from the prompt, we need to adjust the citation numbers after obtaining the inferior answers. Specific examples of citation number adjustments are shown in Appendix~\ref{appendix:Examples}.

\subsubsection{Response Robustness}

Response robustness refers to the ability of the generator to resist interference. The optimization objective from this perspective is to enable the generator to avoid interference from noisy documents. 
% As described in section 2.2, we improve response robustness in situations with low document quality. When constructing preference data, the chosen output is the response that the generator ignores the noisey document and maintains a complete answer, while the rejected output is the response that the generator is influenced by the noisey document, resulting in incomplete or incorrect answers.

\textbf{Noisy document construction:} We categorize noisy documents into two types. The first type is documents related to the question but does not contain the answer, from which we randomly select two documents that do not contain short answers from the top 100 relevant retrieved documents. The second type is irrelevant documents to the question, from which we randomly select two documents that do not contain short answers from the retrieval documents of other questions.

\textbf{Input construction:} The input includes the instruction, the question, and the low-quality prompt documents that mix up to five golden documents and four noisy documents.

\textbf{Chosen output construction:} To simulate a scenario where the generator ignores all noisy documents, we use the chosen output generated in the absence of noisy documents in \S~\ref{sec:RI}. Since there are no noisy documents, we need to adjust the citation numbers. Specific examples of citation number adjustments are shown in Appendix~\ref{appendix:Examples}.

\textbf{Rejected output construction:} Use inputs containing low-quality documents, generate responses with the generator, and filter to retain inferior answers that are incomplete.

\subsubsection{Citation Quality}

Citation quality refers to the generator's ability to cite documents correctly. The optimization objective from this perspective is to enable the generator to cite documents related to the claim correctly and to avoid citing irrelevant documents.
% When constructing preference data, the chosen output is the citation that supports the claim and does not include any irrelevant documents, while the rejected output is the citation that does not support the claim. 

Data construction is divided into two steps: First, we use the generator to generate responses and filter to retain those containing all short answers. Second, we use the \textit{citation rewrite} mechanism to identify incorrect citations that fail the NLI model verification or cite irrelevant documents as rejected output and than correct them as chosen output.

\textbf{Construction of input:} The input includes the instruction, the questions, the prompt documents, and the response up to the incorrect citation.

\textbf{Construction of chosen output:} Correct citation after citation rewrite.

\textbf{Construction of rejected output:} Incorrect citation that fails the NLI model verification or cites irrelevant documents.

\subsubsection{Staged Preference Optimization}

Given the substantial discrepancies in preference information from different perspectives, we divide the preference optimization process into multiple sub-stages to independently optimize the RAG preferences from each perspective. Specifically, the first and second sub-stages optimize response informativeness and response robustness, respectively. The rejected outputs in these stages are generated by the generator that has undergone instruction fine-tuning. The third sub-stage optimizes citation quality, with preference data generated by the generator that has undergone optimization in the first and second sub-stages.
\section{Experiment}
\begin{table*}[t]
\centering
\resizebox{\linewidth}{!}{
\begin{tabular}{lcccccccccccc}
\toprule
\textbf{Datasets} & \multicolumn{3}{c}{\textbf{ASQA}} & 
\multicolumn{3}{c}{\textbf{WebQuestions}}  & 
\multicolumn{3}{c}{\textbf{Natural Questions}}  &
\multicolumn{3}{c}{\textbf{TriviaQA}}\\
 \cmidrule(lr){2-4}  \cmidrule(lr){5-7}  \cmidrule(lr){8-10}  \cmidrule(lr){11-13} 
& {\textbf{Correct}}  & \multicolumn{2}{c}{\textbf{Citation} } &\textbf{Correct} & \multicolumn{2}{c}{\textbf{Citation} } &\textbf{Correct} & \multicolumn{2}{c}{\textbf{Citation} } &\textbf{Correct} & \multicolumn{2}{c}{\textbf{Citation} } \\

\cmidrule(lr){2-2} \cmidrule(lr){3-4}  \cmidrule(lr){5-5}  \cmidrule(lr){6-7}   \cmidrule(lr){8-8} \cmidrule(lr){9-10}   \cmidrule(lr){11-11} \cmidrule(lr){12-13} 

\textbf{Generators} & \textbf{EM} & \textbf{Rec} & \textbf{Prec} & \textbf{EM} & \textbf{Rec} & \textbf{Prec} & \textbf{EM} & \textbf{Rec} & \textbf{Prec} & \textbf{EM} & \textbf{Rec} & \textbf{Prec} \\
\midrule
\headercolor
\multicolumn{13}{c}{\textbf{Base Generators}} \\
\textsc{Llama2-7b-Chat} & 35.61 & 20.51 & 34.51 & 40.91 & 24.36 & 40.27 & 48.82 & 20.06 & 34.71 & 67.69 & 22.81 & 37.27 \\
\textsc{Llama2-13b-Chat} & 27.24 & 16.02 & 24.87 & 25.45 & 13.75 & 21.03 & 34.47 & 14.58 & 22.57 & 54.55 & 15.31 & 23.31   \\
\textsc{Llama3-8b-Instruct} & 34.77 & 50.23 & 51.67 & 39.69 & 49.86 & 51.55 & 49.71 & 47.82 & 49.80 & 66.98 & 45.78 & 45.72  \\
\midrule
\headercolor
\multicolumn{13}{c}{\textbf{Other Training-Needed Methods}} \\
RetRobust-13B & 26.41 & - & - & 33.71 & - & - & 43.08 & - & - & 73.13 & - & -  \\
Self-RAG-13B &  31.61 & 58.77 & 68.88 & 38.64 & 60.46 & 71.75 & 48.09 & 63.65 & 72.75 & 66.95 & 63.62 & 65.24  \\
SFT on Chosen &  34.49 & 72.61 & 69.29 & 39.07 & 80.81 & 77.66 & 50.30 & 73.77 & 70.76 & 64.59 & 66.15 & 62.75  \\
\midrule
\headercolor
\multicolumn{13}{c}{\textbf{\ours{} (Ours)}} \\
\textsc{Llama2-7b}+\ours{} & \textbf{46.16} & \textbf{77.66} & \textbf{76.82} & \textbf{48.37} & \textbf{79.36} & \textbf{78.16} & \textbf{63.27} & \textbf{76.60} & \textbf{76.49} & \textbf{72.52} & \textbf{70.86} & \textbf{69.89}  \\
\textsc{Llama2-13b+\ours{}} & \textbf{46.53} & \textbf{80.68} & \textbf{77.73} & \textbf{48.72} & \textbf{80.90} & \textbf{78.41} & \textbf{62.71} & \textbf{78.72} & \textbf{75.89} & \textbf{73.56} & \textbf{71.96} & \textbf{68.95}  \\
\textsc{Llama3-8b}+\ours{} & \textbf{47.03} & \textbf{81.97} & \textbf{79.93} & \textbf{48.59} & \textbf{84.31} & \textbf{79.84} & \textbf{62.29} & \textbf{81.26} & \textbf{79.08} & \textbf{73.79} & \textbf{74.07} & \textbf{71.05}  \\
\bottomrule
\end{tabular}
}
\caption{Main results of \ours{} and baselines on four QA datasets. The best performances within the same backbone generator are marked in \textbf{bold}.}
\label{tab:main res}
\vspace{-10pt}
\end{table*}
\begin{table*}[t]
\centering
\resizebox{\linewidth}{!}{
\begin{tabular}{lcccccccccccccccc}
\toprule
\textbf{Datasets} & \multicolumn{4}{c}{\textbf{ASQA}} & 
\multicolumn{4}{c}{\textbf{WebQuestions}}  & 
\multicolumn{4}{c}{\textbf{Natural Questions}}  &
\multicolumn{4}{c}{\textbf{TriviaQA}}\\
 \cmidrule(lr){2-5}  \cmidrule(lr){6-9}  \cmidrule(lr){10-13}  \cmidrule(lr){14-17} 
& {\textbf{Correct}}  & \multicolumn{3}{c}{\textbf{Citation} } &\textbf{Correct} & \multicolumn{3}{c}{\textbf{Citation} } &\textbf{Correct} & \multicolumn{3}{c}{\textbf{Citation} } &\textbf{Correct} & \multicolumn{3}{c}{\textbf{Citation} } \\

\cmidrule(lr){2-2} \cmidrule(lr){3-5}  \cmidrule(lr){6-6}  \cmidrule(lr){7-9}   \cmidrule(lr){10-10} \cmidrule(lr){11-13}   \cmidrule(lr){14-14} \cmidrule(lr){15-17} 

\textbf{Generators} & \textbf{EM} & \textbf{Rec} & \textbf{Prec} & \textbf{F1} & \textbf{EM} & \textbf{Rec} & \textbf{Prec} & \textbf{F1} & \textbf{EM} & \textbf{Rec} & \textbf{Prec} & \textbf{F1} & \textbf{EM} & \textbf{Rec} & \textbf{Prec} & \textbf{F1} \\
\midrule
\headercolor
\multicolumn{17}{c}{\textbf{\textsc{Llama2-7b-Chat}}} \\
Base Generator & 35.61 & 20.51 & 34.51 & 25.73 & 40.91 & 24.36 & 40.27 & 30.36 & 48.82 & 20.06 & 34.71 & 25.43 & 67.69 & 22.81 & 37.27 & 28.30 \\
+IFT & 37.98 & 75.69 & 70.17 & 72.83 & 42.76 & 75.85 & 70.50 & 73.08 & 54.59 & 74.34 & 68.95 & 71.54 & 68.51 & 65.41 & 60.34 & 62.77 \\
+IFT+RI & 43.95 & \textbf{77.68} & 68.32 & 72.70 & 47.75 & 78.93 & 69.32 & 73.81 & 60.11 & 76.19 & 66.22 & 70.85 & 71.61 & 69.00 & 60.23 & 64.32 \\
+IFT+RI+RR & \textbf{46.37} & 75.25 & 71.58 & 73.37 & 48.03 & 79.00 & 74.31 & 76.58 & 62.99 & 75.55 & 71.12 & 73.27 & 72.36 & 69.23 & 64.74 & 66.91 \\
+IFT+RI+RR+CQ & 46.16 & 77.66 & \textbf{76.82} & \textbf{77.24} & \textbf{48.37} & \textbf{79.36} & \textbf{78.16} & \textbf{78.76} & \textbf{63.27} & \textbf{76.60} & \textbf{76.49} & \textbf{76.54} & \textbf{72.52} & \textbf{70.86} & \textbf{69.89} & \textbf{70.37} \\
\midrule
\headercolor
\multicolumn{17}{c}{\textbf{\textsc{Llama2-13b-Chat}}} \\
Base Generator & 27.24 & 16.02 & 24.87 & 19.49 & 25.45 & 13.75 & 21.03 & 16.63 & 34.47 & 14.58 & 22.57 & 17.72 & 54.55 & 15.31 & 23.31 & 18.49 \\
+IFT & 38.23 & 76.28 & 70.82 & 73.45 & 42.04 & 78.87 & 73.05 & 75.85 & 54.85 & 75.75 & 69.91 & 72.71 & 69.54 & 66.89 & 61.81 & 64.25 \\
+IFT+RI & 42.46 & 80.24 & 70.05 & 74.80 & 46.66 & \textbf{81.74} & 71.59 & 76.33 & 58.68 & 78.33 & 68.09 & 72.85 & 72.15 & 70.26 & 60.88 & 65.24 \\
+IFT+RI+RR & 45.50 & 77.72 & 75.57 & 76.63 & 48.16 & 80.23 & 77.40 & 78.79 & 61.58 & 76.82 & 74.17 & 75.47 & 72.78 & 68.79 & 66.57 & 67.66 \\
+IFT+RI+RR+CQ & \textbf{46.53} & \textbf{80.68} & \textbf{77.73} & \textbf{79.18} & \textbf{48.72} & 80.90 & \textbf{78.41} & \textbf{79.64} & \textbf{62.71} & \textbf{78.72} & \textbf{75.89} & \textbf{77.28} & \textbf{73.56} & \textbf{71.96} & \textbf{68.95} & \textbf{70.42} \\
\midrule
\headercolor
\multicolumn{17}{c}{\textbf{\textsc{Llama3-8b-Instruct}}} \\
Base Generator & 34.77 & 50.23 & 51.67 & 50.94 & 39.69 & 49.86 & 51.55 & 50.69 & 49.71 & 47.82 & 49.80 & 48.79 & 66.98 & 45.78 & 45.72 & 45.75 \\
+IFT & 38.52 & 76.78 & 71.74 & 74.18 & 42.80 & 79.75 & 74.03 & 76.78 & 54.95 & 81.80 & 71.40 & 76.25 & 69.35 & 69.00 & 63.58 & 66.18 \\
+IFT+RI & 44.16 & \textbf{82.48} & 72.82 & 77.35 & 47.52 & 83.54 & 73.62 & 78.27 & 61.13 & \textbf{82.94} & 76.01 & 79.33 & 72.78 & 73.35 & 63.39 & 68.01 \\
+IFT+RI+RR & \textbf{47.35} & 81.17 & 74.65 & 77.77 & 48.48 & 84.11 & 77.12 & 80.46 & \textbf{62.71} & 82.91 & 75.35 & 78.95 & 73.62 & \textbf{75.14} & 68.18 & 71.49 \\
+IFT+RI+RR+CQ & 47.03 & 81.97 & \textbf{79.93} & \textbf{80.94} & \textbf{48.59} & \textbf{84.31} & \textbf{79.84} & \textbf{82.01} & 62.29 & 81.26 & \textbf{79.08} & \textbf{80.16} & \textbf{73.79} & 74.07 & \textbf{71.05} & \textbf{72.53} \\

\bottomrule
\end{tabular}
}
\caption{Evaluation results for each training phase. IFT stands for instruction fine-tuning, RI denotes the optimization of response informativeness, RR indicates the optimization of response robustness, and CQ represents the optimization of citation quality. The best scores are highlighted in \textbf{bold}.}
\label{tab:stage res}
\vspace{-10pt}
\end{table*}

\subsection{Experimental Setup}

\subsubsection{Dataset and Evaluation Methodology}

\paragraph{Datasets}
The questions used for constructing our training data are sourced from the ASQA \citep{ASQA}, WebQuestions \citep{WebQ}, and Natural Questions \citep{NQ} training splits. Detailed statistics for the training set can be found in Appendix~\ref{appendix: data statistics}. The evaluation data is sourced from the test splits of the three datasets mentioned above, along with TriviaQA \citep{TQA}, which serves as an unseen dataset to assess the \textbf{out-of-domain} generalizability of the generator.

\paragraph{Evaluation Metrics}
To align with the optimization objectives discussed in \S~\ref{sec:Optimization Objectives}, we evaluate the generator's performance regarding correctness and citation quality. Following ALCE \citep{gao2023enabling} and \textsc{Vtg} \citep{sun2023towards}, for correctness, we assess whether the short answers (provided by the dataset) are exact substrings of the generation to calculate the \textbf{exact match} (EM) score. For citation quality, we use \textbf{citation recall} to evaluate whether the output is fully supported by the cited documents and \textbf{citation precision} to assess whether irrelevant documents are cited.

\subsubsection{Implementation Details}
We selected three general LLMs as the base RAG generator: \textsc{Llama2-7b-chat}, \textsc{Llama2-13b-chat}, and \textsc{Llama3-8b-Instruct}.
% For the NLI model in the citation rewrite and citation quality evaluation, we utilized the model TRUE \cite{TRUE}, a \textsc{T5-11B} model fine-tuned on a collection of NLI datasets.
For the NLI model of citation rewrite and citation quality evaluation, we follow the works of \citet{gao2023enabling},  \citet{sun2023towards}, and \citet{huang2024training}, utilizing the NLI model TRUE \cite{TRUE}, a T5-11B model that is fine-tuned on a collection of NLI datasets. Meanwhile, ALCE has verified through extensive human evaluations that this NLI model correlates with human judgment.
%\citet{Self-RAG},
For retrieval, we employed the Wikipedia dump from December 20, 2018, as our retrieval corpus and used GTR \citep{GTR} as our dense retriever.

We performed full fine-tuning on the generators, with the training hyperparameters detailed in Appendix~\ref{appendix:Hyperparameters}. All generators were trained on the same dataset, in which the RAG generator used to generate preference data was the \textsc{Llama2-7b-chat} at different stages of fine-tuning.

\subsubsection{Baselines}
We adopt the following baseline systems.
1) \textbf{Base generator}, the general LLM without further optimization. 
2) \textbf{RetRobust-13B} \citep{RetRobust}, a method that enhances the robustness of the RAG generator through SFT. 
3) \textbf{Self-RAG-13B} \citep{Self-RAG}, a method that improves the performance of the RAG generator through a rank-then-generate pipeline. 
4) \textbf{SFT on chosen}, to compare preference optimization and SFT, we use the chosen output of preference data for SFT on the \textsc{Llama2-7b-chat}. 
All the methods above were evaluated using the same prompt documents used for \ours{} to facilitate a fair comparison.

\subsection{Main Results}
\label{sec: main res}
The evaluation results of \ours{} and the baseline across four QA datasets are presented in Table~\ref{tab:main res}. We have made the following observations: First, \ours{} consistently and significantly improves the performance of various LLMs in RAG scenarios, with an average increase of 13.97\% in EM score, 49.77\% in citation recall, and 39.58\% in citation precision. This demonstrates that our training data is applicable to different LLMs, showcasing strong generalizability. Meanwhile, \ours{} does not compromise the fluency of the generator’s outputs.\footnote{A detailed analysis of response fluency is presented in Appendix~\ref{appendix:fluency}.} Second, \ours{} significantly outperforms the baselines, indicating that compared to end-to-end generators trained via SFT and pipeline generators, the end-to-end generator trained through preference optimization aligns better with RAG preferences, resulting in superior performance.

\subsection{Ablation Study}
The evaluation results for each training phase are presented in Table~\ref{tab:stage res}. We observed that each training phase generally contributes to improvements in both correctness and citation quality, indicating that optimization of RAG requirements in each perspective is necessary. Furthermore, correctness and citation quality are complementary to each other. Detailed Analysis of why optimizing citation quality might have a negative impact on citation recall is presented in Appendix~\ref{appendix: negative recall}.

To more intuitively demonstrate the impact of preference optimization on the generator, we present detailed LLM evaluation in Appendix~\ref{appendix: llm eval} and case studies in Appendix~\ref{appdix:case study}.

\subsection{Impact of Preference Optimization Order}
\begin{table}[t]
\centering
\resizebox{\linewidth}{!}{
\begin{tabular}{lcccc}
\toprule
\textbf{Generators} & \textbf{ASQA} & 
\textbf{WebQ}  & 
\textbf{NQ}  &
\textbf{TQA}\\
\toprule
\textsc{Llama-7b-Chat}+IFT & 37.98 & 42.76 & 54.59 & 68.51 \\
+RI & 43.95 & 47.75 & 60.11 & 71.61 \\
+RR & 25.91 & 32.23 & 52.52 & 37.52 \\
+RR+RI & 44.74 & 47.09 & 61.78 & 71.28 \\
+Mix RI\&RR & 33.72 & 37.84 & 61.43 & 45.66 \\
+RI+RR & \textbf{46.37} & \textbf{48.03} & \textbf{62.99} & \textbf{72.36} \\
\bottomrule
\end{tabular}
}
\caption{EM score results of different preference optimization orders. Best scores are highlighted in \textbf{bold}.}
\label{tab:PO order}
\vspace{-10pt}
\end{table}
In the main experiment, we first optimized response informativeness, followed by response robustness. We further investigated the impact of altering the optimization order on performance, with results shown in Table~\ref{tab:PO order}. Our findings are as follows: First, optimizing response robustness before response informativeness resulted in lower performance than the main experimental setup. Second, skipping informativeness and directly optimizing robustness or merging the data to optimize simultaneously can lead to a decline in performance. We believe this is because optimizing informativeness enables the generator to learn how to effectively utilize retrieved documents, which is the fundamental capability in the RAG system. In contrast, optimizing robustness requires the generator to learn to reject irrelevant documents, which is a more advanced skill.\footnote{Detailed analysis of why rejecting irrelevant documents is a more advanced skill is showcased in Appendix~\ref{appendix: detailed analysis}.} Learning skills at different levels requires a reasonable arrangement in learning order, which is similar to curriculum learning \citep{bengio2009curriculum}. Neglecting the foundational skills or mixing different skill levels during learning may hurt the model's original capabilities.

\subsection{Further Comparison of DPO and SFT}
% \subsection{Further Comparison of DPO and SFT under RAG Preference Alignment}
\begin{table}[t]
\centering
\resizebox{\linewidth}{!}{
\begin{tabular}{lcccc}
\toprule
\textbf{Generators} & \textbf{ASQA} & 
\textbf{WebQ}  & 
\textbf{NQ}  &
\textbf{TQA}\\
\toprule
\textsc{Llama-7b-Chat}+IFT & 37.98 & 42.76 & 54.59 & 68.51 \\
+SFT RI(Chosen) & 38.93 & 43.69 & 54.84 & 69.10 \\
+SFT RI+RR(Chosen) & 32.17 & 37.02 & 47.01 & 62.56 \\
+DPO RI & 43.95 & 47.75 & 60.11 & 71.61 \\
+DPO RI+RR & 46.37 & 48.03 & 62.99 & 72.36 \\
\bottomrule
\end{tabular}
}
\caption{Further comparison of DPO and SFT under response informativeness (RI) and response robustness (RR). Results of the EM score.}
\label{tab:DPO vs SFT}
\vspace{-10pt}
\end{table}
\label{sec: dpo vs sft}

As mentioned in \S~\ref{sec 2.2}, there is a significant disparity in the optimization directions of response informativeness and robustness. We further explore the differences between DPO and SFT when optimizing these two directions sequentially. As shown in Table~\ref{tab:DPO vs SFT}, we observe that, when using SFT for training, optimizing response informativeness first can enhance the generator's performance. However, subsequent optimization for response robustness leato a significant performance decline, exposing SFT's vulnerability to catastrophic forgetting \citep{french1999catastrophic} when undergoing two relatively different optimizations consecutively. In contrast, DPO handles this situation well, consistently improving the generator's performance.
\section{Related Work}

\subsection{Retrieval-Augmented Generator}

To satisfy the requirements of RAG system, existing retrieval-augmented generators are primarily categorized into end-to-end and pipeline frameworks. In end-to-end frameworks, models such as RetRobust \citep{yoran2023making} and RAAT \citep{fang2024enhancing} enhance the robustness of the generator by performing SFT on high-quality data.
% Additionally, InFO-RAG \citep{dong2024unsupervised} utilizes unsupervised learning to improve the accuracy and completeness of their outputs. 
In pipeline frameworks, models like DPA-RAG \citep{dong2024understand}, REAR \citep{wang2024rear}, Self-RAG \citep{Self-RAG}, and RankRAG \citep{yu2024rankrag} enhance the robustness of the generator by explicitly re-ranking the retrieved documents. Furthermore, \textsc{Vtg} \citep{sun2023towards} improves citation quality by introducing explicit citation verification and modification. However, these methods still struggle to align the generator with the RAG requirements fully.

\subsection{Fine-tuning Approaches for LMs}

The mainstream approaches for fine-tuning language models in generative tasks include SFT \citep{GPT} and reinforcement learning from human feedback (RLHF) \citep{RLHF}. SFT involves constructing input-output pairs to teach the model how to complete tasks. RLHF methods like DPO \citep{DPO} incorporate training data that includes both superior and inferior outputs, allowing the model to learn preference information. RLHF is more effective than SFT in aligning the model with task preferences.
\section{Conclusion}

In this work, we propose \ours{}, a method for optimizing the generator of RAG systems to align with specific RAG requirements comprehensively. The training process includes instruction fine-tuning and multi-perspective preference optimization. We conducted extensive experiments on four QA benchmarks and three LLMs, demonstrating that \ours{} can significantly enhance the generator's response informativeness, response robustness, and citation quality.
\section*{Limitations}
Our method requires fine-tuning in four stages, including one instruction fine-tuning stage and three preference optimization stages. This results in a cumbersome search for the optimal hyperparameter settings during training. We have presented the hyperparameter details in Appendix~\ref{appendix:Hyperparameters} to prevent other researchers from duplicating the search for the best hyperparameter settings.
\section*{Ethics Statement}
This work was conducted in strict compliance with the ACL Ethics Policy. 
All datasets and models used for experiment are publicly available. 
Furthermore, our work aims to explore a RAG generator training method. 
We do not foresee any negative ethical impacts arising from our work.

% Bibliography entries for the entire Anthology, followed by custom entries
%\bibliography{anthology,custom}
% Custom bibliography entries only
\bibliography{custom}

\begin{thebibliography}{28}
\providecommand{\natexlab}[1]{#1}

\bibitem[{Asai et~al.(2024)Asai, Wu, Wang, Sil, and Hajishirzi}]{Self-RAG}
Akari Asai, Zeqiu Wu, Yizhong Wang, Avirup Sil, and Hannaneh Hajishirzi. 2024.
\newblock \href {https://openreview.net/forum?id=hSyW5go0v8} {Self-rag: Learning to retrieve, generate, and critique through self-reflection}.
\newblock In \emph{The Twelfth International Conference on Learning Representations, {ICLR} 2024, Vienna, Austria, May 7-11, 2024}. OpenReview.net.

\bibitem[{Bengio et~al.(2009)Bengio, Louradour, Collobert, and Weston}]{bengio2009curriculum}
Yoshua Bengio, J{\'e}r{\^o}me Louradour, Ronan Collobert, and Jason Weston. 2009.
\newblock Curriculum learning.
\newblock In \emph{Proceedings of the 26th annual international conference on machine learning}, pages 41--48.

\bibitem[{Berant et~al.(2013)Berant, Chou, Frostig, and Liang}]{WebQ}
Jonathan Berant, Andrew Chou, Roy Frostig, and Percy Liang. 2013.
\newblock \href {https://aclanthology.org/D13-1160/} {Semantic parsing on freebase from question-answer pairs}.
\newblock In \emph{Proceedings of the 2013 Conference on Empirical Methods in Natural Language Processing, {EMNLP} 2013, 18-21 October 2013, Grand Hyatt Seattle, Seattle, Washington, USA, {A} meeting of SIGDAT, a Special Interest Group of the {ACL}}, pages 1533--1544. {ACL}.

\bibitem[{Christiano et~al.(2017)Christiano, Leike, Brown, Martic, Legg, and Amodei}]{RLHF}
Paul~F Christiano, Jan Leike, Tom Brown, Miljan Martic, Shane Legg, and Dario Amodei. 2017.
\newblock Deep reinforcement learning from human preferences.
\newblock \emph{Advances in neural information processing systems}, 30.

\bibitem[{Dong et~al.(2024{\natexlab{a}})Dong, Zhu, Zhang, Wang, Dou, and Wen}]{dong2024understand}
Guanting Dong, Yutao Zhu, Chenghao Zhang, Zechen Wang, Zhicheng Dou, and Ji{-}Rong Wen. 2024{\natexlab{a}}.
\newblock \href {https://doi.org/10.48550/ARXIV.2406.18676} {Understand what {LLM} needs: Dual preference alignment for retrieval-augmented generation}.
\newblock \emph{CoRR}, abs/2406.18676.

\bibitem[{Dong et~al.(2024{\natexlab{b}})Dong, Liu, Ai, Wu, Li, Liu, Wang, Yin, and Ma}]{dong2024unsupervised}
Qian Dong, Yiding Liu, Qingyao Ai, Zhijing Wu, Haitao Li, Yiqun Liu, Shuaiqiang Wang, Dawei Yin, and Shaoping Ma. 2024{\natexlab{b}}.
\newblock \href {https://doi.org/10.1145/3626772.3657689} {Unsupervised large language model alignment for information retrieval via contrastive feedback}.
\newblock In \emph{Proceedings of the 47th International {ACM} {SIGIR} Conference on Research and Development in Information Retrieval, {SIGIR} 2024, Washington DC, USA, July 14-18, 2024}, pages 48--58. {ACM}.

\bibitem[{Fang et~al.(2024)Fang, Bai, Ni, Yang, Chen, and Xu}]{fang2024enhancing}
Feiteng Fang, Yuelin Bai, Shiwen Ni, Min Yang, Xiaojun Chen, and Ruifeng Xu. 2024.
\newblock \href {https://aclanthology.org/2024.acl-long.540} {Enhancing noise robustness of retrieval-augmented language models with adaptive adversarial training}.
\newblock In \emph{Proceedings of the 62nd Annual Meeting of the Association for Computational Linguistics (Volume 1: Long Papers), {ACL} 2024, Bangkok, Thailand, August 11-16, 2024}, pages 10028--10039. Association for Computational Linguistics.

\bibitem[{French and Chater(2002)}]{french1999catastrophic}
Robert~M. French and Nick Chater. 2002.
\newblock \href {https://doi.org/10.1162/08997660260028700} {Using noise to compute error surfaces in connectionist networks: {A} novel means of reducing catastrophic forgetting}.
\newblock \emph{Neural Comput.}, 14(7):1755--1769.

\bibitem[{Gao et~al.(2023{\natexlab{a}})Gao, Yen, Yu, and Chen}]{gao2023enabling}
Tianyu Gao, Howard Yen, Jiatong Yu, and Danqi Chen. 2023{\natexlab{a}}.
\newblock \href {https://doi.org/10.18653/V1/2023.EMNLP-MAIN.398} {Enabling large language models to generate text with citations}.
\newblock In \emph{Proceedings of the 2023 Conference on Empirical Methods in Natural Language Processing, {EMNLP} 2023, Singapore, December 6-10, 2023}, pages 6465--6488. Association for Computational Linguistics.

\bibitem[{Gao et~al.(2023{\natexlab{b}})Gao, Xiong, Gao, Jia, Pan, Bi, Dai, Sun, Guo, Wang, and Wang}]{gao2023retrieval}
Yunfan Gao, Yun Xiong, Xinyu Gao, Kangxiang Jia, Jinliu Pan, Yuxi Bi, Yi~Dai, Jiawei Sun, Qianyu Guo, Meng Wang, and Haofen Wang. 2023{\natexlab{b}}.
\newblock \href {https://doi.org/10.48550/ARXIV.2312.10997} {Retrieval-augmented generation for large language models: {A} survey}.
\newblock \emph{CoRR}, abs/2312.10997.

\bibitem[{Honovich et~al.(2022)Honovich, Aharoni, Herzig, Taitelbaum, Kukliansky, Cohen, Scialom, Szpektor, Hassidim, and Matias}]{TRUE}
Or~Honovich, Roee Aharoni, Jonathan Herzig, Hagai Taitelbaum, Doron Kukliansky, Vered Cohen, Thomas Scialom, Idan Szpektor, Avinatan Hassidim, and Yossi Matias. 2022.
\newblock \href {https://doi.org/10.18653/V1/2022.NAACL-MAIN.287} {{TRUE:} re-evaluating factual consistency evaluation}.
\newblock In \emph{Proceedings of the 2022 Conference of the North American Chapter of the Association for Computational Linguistics: Human Language Technologies, {NAACL} 2022, Seattle, WA, United States, July 10-15, 2022}, pages 3905--3920. Association for Computational Linguistics.

\bibitem[{Huang et~al.(2024)Huang, Wu, Hu, and Wang}]{huang2024training}
Chengyu Huang, Zeqiu Wu, Yushi Hu, and Wenya Wang. 2024.
\newblock \href {https://doi.org/10.18653/V1/2024.ACL-LONG.161} {Training language models to generate text with citations via fine-grained rewards}.
\newblock In \emph{Proceedings of the 62nd Annual Meeting of the Association for Computational Linguistics (Volume 1: Long Papers), {ACL} 2024, Bangkok, Thailand, August 11-16, 2024}, pages 2926--2949. Association for Computational Linguistics.

\bibitem[{Joshi et~al.(2017)Joshi, Choi, Weld, and Zettlemoyer}]{TQA}
Mandar Joshi, Eunsol Choi, Daniel~S. Weld, and Luke Zettlemoyer. 2017.
\newblock \href {https://doi.org/10.18653/V1/P17-1147} {Triviaqa: {A} large scale distantly supervised challenge dataset for reading comprehension}.
\newblock In \emph{Proceedings of the 55th Annual Meeting of the Association for Computational Linguistics, {ACL} 2017, Vancouver, Canada, July 30 - August 4, Volume 1: Long Papers}, pages 1601--1611. Association for Computational Linguistics.

\bibitem[{Kwiatkowski et~al.(2019)Kwiatkowski, Palomaki, Redfield, Collins, Parikh, Alberti, Epstein, Polosukhin, Devlin, Lee, Toutanova, Jones, Kelcey, Chang, Dai, Uszkoreit, Le, and Petrov}]{NQ}
Tom Kwiatkowski, Jennimaria Palomaki, Olivia Redfield, Michael Collins, Ankur~P. Parikh, Chris Alberti, Danielle Epstein, Illia Polosukhin, Jacob Devlin, Kenton Lee, Kristina Toutanova, Llion Jones, Matthew Kelcey, Ming{-}Wei Chang, Andrew~M. Dai, Jakob Uszkoreit, Quoc Le, and Slav Petrov. 2019.
\newblock \href {https://doi.org/10.1162/TACL\_A\_00276} {Natural questions: a benchmark for question answering research}.
\newblock \emph{Trans. Assoc. Comput. Linguistics}, 7:452--466.

\bibitem[{Li et~al.(2023)Li, Zhang, Dubois, Taori, Gulrajani, Guestrin, Liang, and Hashimoto}]{alpaca_eval}
Xuechen Li, Tianyi Zhang, Yann Dubois, Rohan Taori, Ishaan Gulrajani, Carlos Guestrin, Percy Liang, and Tatsunori~B. Hashimoto. 2023.
\newblock Alpacaeval: An automatic evaluator of instruction-following models.
\newblock \url{https://github.com/tatsu-lab/alpaca_eval}.

\bibitem[{Ni et~al.(2022)Ni, Qu, Lu, Dai, {\'{A}}brego, Ma, Zhao, Luan, Hall, Chang, and Yang}]{GTR}
Jianmo Ni, Chen Qu, Jing Lu, Zhuyun Dai, Gustavo~Hern{\'{a}}ndez {\'{A}}brego, Ji~Ma, Vincent~Y. Zhao, Yi~Luan, Keith~B. Hall, Ming{-}Wei Chang, and Yinfei Yang. 2022.
\newblock \href {https://doi.org/10.18653/V1/2022.EMNLP-MAIN.669} {Large dual encoders are generalizable retrievers}.
\newblock In \emph{Proceedings of the 2022 Conference on Empirical Methods in Natural Language Processing, {EMNLP} 2022, Abu Dhabi, United Arab Emirates, December 7-11, 2022}, pages 9844--9855. Association for Computational Linguistics.

\bibitem[{Pillutla et~al.(2021)Pillutla, Swayamdipta, Zellers, Thickstun, Welleck, Choi, and Harchaoui}]{MAUVE}
Krishna Pillutla, Swabha Swayamdipta, Rowan Zellers, John Thickstun, Sean Welleck, Yejin Choi, and Za{\"{\i}}d Harchaoui. 2021.
\newblock \href {https://proceedings.neurips.cc/paper/2021/hash/260c2432a0eecc28ce03c10dadc078a4-Abstract.html} {{MAUVE:} measuring the gap between neural text and human text using divergence frontiers}.
\newblock In \emph{Advances in Neural Information Processing Systems 34: Annual Conference on Neural Information Processing Systems 2021, NeurIPS 2021, December 6-14, 2021, virtual}, pages 4816--4828.

\bibitem[{Radford(2018)}]{GPT}
Alec Radford. 2018.
\newblock Improving language understanding by generative pre-training.

\bibitem[{Rafailov et~al.(2023)Rafailov, Sharma, Mitchell, Manning, Ermon, and Finn}]{DPO}
Rafael Rafailov, Archit Sharma, Eric Mitchell, Christopher~D. Manning, Stefano Ermon, and Chelsea Finn. 2023.
\newblock \href {http://papers.nips.cc/paper\_files/paper/2023/hash/a85b405ed65c6477a4fe8302b5e06ce7-Abstract-Conference.html} {Direct preference optimization: Your language model is secretly a reward model}.
\newblock In \emph{Advances in Neural Information Processing Systems 36: Annual Conference on Neural Information Processing Systems 2023, NeurIPS 2023, New Orleans, LA, USA, December 10 - 16, 2023}.

\bibitem[{Stelmakh et~al.(2022)Stelmakh, Luan, Dhingra, and Chang}]{ASQA}
Ivan Stelmakh, Yi~Luan, Bhuwan Dhingra, and Ming{-}Wei Chang. 2022.
\newblock \href {https://doi.org/10.18653/V1/2022.EMNLP-MAIN.566} {{ASQA:} factoid questions meet long-form answers}.
\newblock In \emph{Proceedings of the 2022 Conference on Empirical Methods in Natural Language Processing, {EMNLP} 2022, Abu Dhabi, United Arab Emirates, December 7-11, 2022}, pages 8273--8288. Association for Computational Linguistics.

\bibitem[{Sun et~al.(2023)Sun, Cai, Wang, Hou, Wei, Wang, Zhang, and Yin}]{sun2023towards}
Hao Sun, Hengyi Cai, Bo~Wang, Yingyan Hou, Xiaochi Wei, Shuaiqiang Wang, Yan Zhang, and Dawei Yin. 2023.
\newblock Towards verifiable text generation with evolving memory and self-reflection.
\newblock \emph{arXiv preprint arXiv:2312.09075}.

\bibitem[{Tonmoy et~al.(2024)Tonmoy, Zaman, Jain, Rani, Rawte, Chadha, and Das}]{Hallucination_survey}
S.~M. Towhidul~Islam Tonmoy, S.~M.~Mehedi Zaman, Vinija Jain, Anku Rani, Vipula Rawte, Aman Chadha, and Amitava Das. 2024.
\newblock \href {https://doi.org/10.48550/ARXIV.2401.01313} {A comprehensive survey of hallucination mitigation techniques in large language models}.
\newblock \emph{CoRR}, abs/2401.01313.

\bibitem[{Turpin et~al.(2023)Turpin, Michael, Perez, and Bowman}]{turpin2024language}
Miles Turpin, Julian Michael, Ethan Perez, and Samuel~R. Bowman. 2023.
\newblock \href {http://papers.nips.cc/paper\_files/paper/2023/hash/ed3fea9033a80fea1376299fa7863f4a-Abstract-Conference.html} {Language models don't always say what they think: Unfaithful explanations in chain-of-thought prompting}.
\newblock In \emph{Advances in Neural Information Processing Systems 36: Annual Conference on Neural Information Processing Systems 2023, NeurIPS 2023, New Orleans, LA, USA, December 10 - 16, 2023}.

\bibitem[{Wang et~al.(2024)Wang, Ren, Li, Zhao, Liu, and Wen}]{wang2024rear}
Yuhao Wang, Ruiyang Ren, Junyi Li, Wayne~Xin Zhao, Jing Liu, and Ji-Rong Wen. 2024.
\newblock Rear: A relevance-aware retrieval-augmented framework for open-domain question answering.
\newblock \emph{arXiv preprint arXiv:2402.17497}.

\bibitem[{Yoran et~al.(2024{\natexlab{a}})Yoran, Wolfson, Ram, and Berant}]{yoran2023making}
Ori Yoran, Tomer Wolfson, Ori Ram, and Jonathan Berant. 2024{\natexlab{a}}.
\newblock \href {https://openreview.net/forum?id=ZS4m74kZpH} {Making retrieval-augmented language models robust to irrelevant context}.
\newblock In \emph{The Twelfth International Conference on Learning Representations, {ICLR} 2024, Vienna, Austria, May 7-11, 2024}. OpenReview.net.

\bibitem[{Yoran et~al.(2024{\natexlab{b}})Yoran, Wolfson, Ram, and Berant}]{RetRobust}
Ori Yoran, Tomer Wolfson, Ori Ram, and Jonathan Berant. 2024{\natexlab{b}}.
\newblock \href {https://openreview.net/forum?id=ZS4m74kZpH} {Making retrieval-augmented language models robust to irrelevant context}.
\newblock In \emph{The Twelfth International Conference on Learning Representations, {ICLR} 2024, Vienna, Austria, May 7-11, 2024}. OpenReview.net.

\bibitem[{Yu et~al.(2024)Yu, Ping, Liu, Wang, You, Zhang, Shoeybi, and Catanzaro}]{yu2024rankrag}
Yue Yu, Wei Ping, Zihan Liu, Boxin Wang, Jiaxuan You, Chao Zhang, Mohammad Shoeybi, and Bryan Catanzaro. 2024.
\newblock Rankrag: Unifying context ranking with retrieval-augmented generation in llms.
\newblock \emph{arXiv preprint arXiv:2407.02485}.

\bibitem[{Zhu et~al.(2024)Zhu, Yan, Shi, Yin, and Sha}]{zhu2024atm}
Junda Zhu, Lingyong Yan, Haibo Shi, Dawei Yin, and Lei Sha. 2024.
\newblock Atm: Adversarial tuning multi-agent system makes a robust retrieval-augmented generator.
\newblock \emph{arXiv preprint arXiv:2405.18111}.

\end{thebibliography}

\appendix

\section{Training Hyperparameters}
\label{appendix:Hyperparameters}

We utilized full fine-tuning for all training stages and employed the same hyperparameter settings for all models.

During the instruction fine-tuning phase, we set the batch size to 128, the learning rate to 2e-5, and trained for one epoch.

In the preference optimization phase, we set the batch size to 64 and trained for one epoch for all stages. For the optimization stages of response informativeness and response robustness, the learning rate is 2e-6. In the citation quality optimization stage, the learning rate is 2e-7.

\section{Compute Resources}
Our experiments were conducted on a server equipped with 1TB of memory and 4 Nvidia A800 80G GPUs.

\section{Training Data Statisics}
\label{appendix: data statistics}
\begin{table}[h]
\centering
\resizebox{0.9\linewidth}{!}{
\begin{tabular}{lcccc}
\toprule
 & \textbf{IFT} & 
\textbf{RI}  & 
\textbf{RR}  &
\textbf{CQ}\\
\toprule
ASQA & 1,714 & 1,046 & 962 & 631 \\
WebQ & 1,681 & 326 & 357 & 653 \\
NQ & 55,463 & 10,416 & 12,080 & 21,241 \\
Sum & 58,858 & 11,788 & 13,399 & 22,525 \\
\bottomrule
\end{tabular}
}
\caption{Training data statistics}
\label{tab:train data statistics}
\end{table}
Detailed information on the data size and composition for each training stage can be found in Table~\ref{tab:train data statistics}.

\subsection{Distribution of Citation Complexity in Instruction Fine-tuning}

In the instruction fine-tuning training data, the distribution of the number of citations per claim is shown in Table~\ref{tab:citations distribution}.

\begin{table}[h]
\centering
\resizebox{\linewidth}{!}{
\begin{tabular}{lccccc}
\toprule
& \textbf{1}  & 
\textbf{2}  &
\textbf{3} &
\textbf{4} &
\textbf{5}\\
\toprule
ASQA & 70.98\% & 24.17\% & 4.24\% & 0.46\% & 0.05\% \\
WebQ & 71.43\% & 21.54\% & 6.26\% & 0.65\% & 0.05\% \\
NQ & 71.87\% & 21.44\% & 6.00\% & 0.52\% & 0.16\% \\
\bottomrule
\end{tabular}
}
\caption{Distribution of Citation Complexity}
\label{tab:citations distribution}
\end{table}

\subsection{The Disparity Between “Chosen” and “Rejected” in Preference Data.}

\begin{table}[h]
\centering
\resizebox{0.88\linewidth}{!}{
\begin{tabular}{lcccc}
\toprule
\multirow{2}{*}{\textbf{Dataset}}& \multirow{2}{*}{\textbf{Preference}} & \textbf{Correct}  & 
\multicolumn{2}{c}{\textbf{Citation}} \\
\cmidrule(lr){3-3} 
\cmidrule(lr){4-5}
& & \textbf{EM} & \textbf{Rec} & \textbf{Prec} \\
\midrule
\headercolor
\multicolumn{5}{c}{\textbf{Response Informativeness}} \\
\multirow{2}{*}{ASQA} & Chosen & 100 & 100 & 100 \\
& Rejected & 38.33 & 77.59 & 75.51 \\
\midrule
\multirow{2}{*}{WebQ} & Chosen & 100 & 100 & 100 \\
& Rejected & 21.14 & 72.90 & 71.00 \\
\midrule
\multirow{2}{*}{NQ} & Chosen & 100 & 100 & 100 \\
& Rejected & 9.07 & 70.53 & 69.62 \\
\midrule
\headercolor
\multicolumn{5}{c}{\textbf{Response Robustness}} \\
\multirow{2}{*}{ASQA} & Chosen & 100 & 100 & 100 \\
& Rejected & 37.30 & 76.62 & 69.19 \\
\midrule
\multirow{2}{*}{WebQ} & Chosen & 100 & 100 & 100 \\
& Rejected & 21.14 & 74.71 & 67.51 \\
\midrule
\multirow{2}{*}{NQ} & Chosen & 100 & 100 & 100 \\
& Rejected & 6.08 & 73.19 & 67.40 \\
\midrule
\headercolor
\multicolumn{5}{c}{\textbf{Citation Quality}} \\
\multirow{2}{*}{ASQA} & Chosen & - & 100 & 100 \\
& Rejected & - & 27.42 & 12.13 \\
\midrule
\multirow{2}{*}{WebQ} & Chosen & - & 100 & 100 \\
& Rejected & - & 25.88 & 13.17 \\
\midrule
\multirow{2}{*}{NQ} & Chosen & - & 100 & 100 \\
& Rejected & - & 27.15 & 13.13 \\

\bottomrule
\end{tabular}
}
\caption{Preference disparity within the preference data}
\label{tab:disparity}
\end{table}

As shown in Table~\ref{tab:disparity}, the disparity in correctness and citation quality between the “Chosen” and “Rejected” in the preference data are presented. We can observe that the preference differences within the preference data are substantial enough to provide relatively rich preference information.

\section{Prompt}

\label{appendix:prompt}
\begin{tcolorbox}[title=Prompt for ChatGPT-3.5]
\small
\textbf{Instruction:} Write an accurate, engaging, and concise answer for the given question using only the provided search results (some of which might be irrelevant) and cite them properly. Use an unbiased and journalistic tone. Always cite for any factual claim. When citing several search results, use [1][2][3]. Cite at least one document and at most three documents in each sentence. If multiple documents support the sentence, only cite a minimum sufficient subset of the documents. \\
\textbf{Qustion:} \{Question\}\\ 
The final answer should contain the following short answers: \{Short answers\} \\
\textbf{Documents:} \{Documents\} \\
\textbf{Answer:}
\end{tcolorbox}

\section{Examples of Adjusting the Citation Number}
\label{appendix:Examples}
\subsection{Rejected Output for Response Informativeness}
When constructing the DPO training data for response informativeness, a question and five golden documents were given, numbered \texttt{[1][2][3][4][5]}. When constructing the rejected output, we would remove some golden documents, such as documents \texttt{[1]} and \texttt{[3]}, leaving documents \texttt{[2]}, \texttt{[4]}, and \texttt{[5]}. To maintain a unified input format, the documents will be renumbered starting from 1. The mapping of the new and old document numbers is shown as Table~\ref{tab: map1}.

\begin{table}[h]
\centering
\resizebox{0.4\linewidth}{!}{
\begin{tabular}{cc}
\toprule
\textbf{Old ID} & 
\textbf{New ID}\\
\toprule
{[2]} & {[1]}\\
{[4]} & {[2]}\\
{[5]} & {[3]}\\
\bottomrule
\end{tabular}
}
\caption{}
\label{tab: map1}
\end{table}

The generator outputs an answer based on the question and the remaining documents: \texttt{Claim1[2]. Claim2[1][3].}

Since the citation numbers in the answer generated by the generator are the new document numbers, and the documents in the input part of the DPO training data are the old numbers, we need to revert the new numbers to the old ones using to the table above. The restored answer is as follows: \texttt{Claim1[4]. Claim2[2][5].}

\subsection{Chosen Output for Response Robustness}
When constructing the training data for response robustness, given a question and documents numbered \texttt{[1][2][3][4][5][6][7][8]}, where \texttt{[1][3][4][7]} are golden documents and \texttt{[2][5][6][8]} are noisy documents. When constructing the chosen output, we will remove all the noisy documents. The mapping of the new and old document numbers is shown as Table~\ref{tab: map2}.

\begin{table}[h]
\centering
\resizebox{0.4\linewidth}{!}{
\begin{tabular}{cc}
\toprule
\textbf{Old ID} & 
\textbf{New ID}\\
\toprule
{[1]} & {[1]}\\
{[3]} & {[2]}\\
{[4]} & {[3]}\\
{[7]} & {[4]}\\
\bottomrule
\end{tabular}
}
\caption{}
\label{tab: map2}
\end{table}

ChatGPT outputs an answer based on the question and the remaining documents: \texttt{Claim1[1][2]. Claim2[3][4].}

Restore the new numbers to the old ones according to the table above. The restored answer is as follows: \texttt{Claim1[1][3]. Claim2[4][7].}

\section{Detailed Analysis of Why Rejecting Irrelevant Documents is a More Advanced Skill}
\label{appendix: detailed analysis}
As shown in Table~\ref{tab:PO order}, in our experiments, skipping the training for informativeness (learning to utilize relevant documents) and directly conducting the training for robustness (learning to ignore irrelevant documents) leads to a decline in model performance. This is because the model may not realize that the documents it needs to ignore are the irrelevant ones, resulting in indiscriminate neglect of most documents. In contrast, performing training for informativeness first and then training for robustness improves performance. This is because the model has already developed a basic ability to utilize documents and can recognize that it should ignore only the irrelevant documents. Therefore, ignoring irrelevant documents is a more advanced skill than fully utilizing every relevant document. If the results were the opposite, it would be counterintuitive and inexplicable.
\section{Detailed Analysis of Why Optimizing Citation Quality Might Have a Negative Impact on citation Recall.}
\label{appendix: negative recall}
It is challenging to simultaneously improve both the precision and recall of citation quality. While enhancing recall, the model may become stricter in assessing the relevance of documents, which can lead to a decrease in precision. Therefore, we introduced the F1 score to evaluate citation quality from a more balanced perspective. Our experimental results indicate that optimizing response informativeness and response robustness can enhance answer correctness, while optimizing citation quality can improve the citation F1 score.
\section{Further analysis of whether PA-RAG affects the fluency of responses}
\label{appendix:fluency}
We follow ALCE \citep{gao2023enabling}, using MAUVE \citep{MAUVE} to evaluate the fluency of the model’s responses. The fluency of the text output by models at various training stages on the ASQA dataset is shown in the Table~\ref{tab:MAUVE}:

\begin{table}[h]
\centering
\resizebox{\linewidth}{!}{
\begin{tabular}{lccc}
\toprule
\textbf{MAUVE score} & \textbf{\textsc{Llama2-7b}} & 
\textbf{\textsc{Llama2-13b}}  & 
\textbf{\textsc{Llama3-8b}}\\
\toprule
Base Generator & 80.10 & 72.40 & 82.30 \\
Base+IFT & 84.99 & 84.00 & 80.86 \\
Base+IFT+RI & 83.88 & 81.41 & 85.87 \\
Base+IFT+RI+RR & 84.77 & 80.72 & 80.73 \\
Base+IFT+RI+RR+CQ & 85.42 & 82.20 & 83.21 \\
\bottomrule
\end{tabular}
}
\caption{The results of the MAUVE score for each training stages. IFT stands for instruction fine-tuning, RI denotes the optimization of response informativeness, RR indicates the optimization of response robustness, and CQ represents the optimization of citation quality.}
\label{tab:MAUVE}
\end{table}
We found that PA-RAG does not affect the fluency of the model’s output.
\section{Specific Criteria and Examples of Inferior Answers that are Incomplete}
\label{appendix:incomplete ans}
Inferior answers refer to responses that do not meet the criteria for correct answers. The criteria for the correct answers are shown in \S~\ref{sec:Optimization Objectives}. Transform these into criteria for inferior answers is as follows:
Let $G$ represent the generator, $x$ denotes the input, $D=\{d_1,d_2,...,d_n\}$ represent the documents retrieved by the retriever, $A=\{a_1,a_2,...,a_n\}$ represent the short answers contained in the documents, and $y$ denote the response generated by the generator. The correct response by the generator can be expressed as:
\begin{equation}
y = G(x, D), \exists a_i \in A, C(y, a_i) = \text{False}
\end{equation}
where $C(y, a_i) = \text{False}$ implies that the answer $a_i$ is not included in $y$.

Specific examples of inferior answers are as follows:

\begin{tcolorbox}[title=Prompt for ChatGPT-3.5]
\small
\textbf{Question:} What is the name of the pirate in SpongeBob? \\
\textbf{Golden short answers:} ['Patchy', 'Burger Beard'] \\
\textbf{Inferior long answer:} The pirate in SpongeBob is named Patchy the Pirate[1]. Patchy the Pirate is the president of the SpongeBob SquarePants fan club and is portrayed as a tyrannical pirate in the plot of the SpongeBob SquarePants film series[1][2].
\end{tcolorbox}

In this inferior answer, the answer ``Burger Beard'' is missing. We determine whether an answer is inferior by checking if any golden short answer is not included in the long answer.
\section{LLM Evaluation}
\label{appendix: llm eval}
\begin{table}[h]
\centering
\resizebox{\linewidth}{!}{
\begin{tabular}{lcccc}
\toprule
\textbf{PA-RAG v.s. IFT Only} & \textbf{ASQA} & 
\textbf{WebQ}  & 
\textbf{NQ}  &
\textbf{TQA}\\
\toprule
\textsc{Llama2-7b-Chat} & 65.30 & 65.11 & 67.09 & 69.39 \\
\textsc{Llama2-13b-Chat} & 73.10 & 70.72 & 72.39 & 74.78 \\
\textsc{Llama3-8b-Instruct} & 71.41 & 66.98 & 69.60 & 71.09 \\
\bottomrule
\end{tabular}
}
\caption{The results of the win rate after preference alignment (PA-RAG) v.s. before preference alignment (instruction fine-tuning only). A higher success rate denotes a larger improvement in response informativeness and response robustness attributed to preference alignment.}
\label{tab:llm eval}
\end{table}

As an accuracy evaluation metric, Exact Match (EM) cannot directly reflect the response informativeness and response robustness. We follow AlpacaEval \citep{alpaca_eval} and use the LLM (\texttt{GPT-3.5-Turbo}) evaluation to qualitatively judge whether the response informativeness and response robustness increase after RAG preference alignment. The prompt for the LLM judgement is as follows:

\begin{tcolorbox}[title=Prompt for ChatGPT-3.5]
\small
I want you to create a leaderboard of different of large-language models. To do so, I will give you the instructions (prompts) given to the models, and the responses of two models. Please rank the models according to the correctness of their answers and whether they make use of valuable documents and avoid being interfered with by irrelevant documents. All inputs and outputs should be python dictionaries.

Here is the prompt: \{instruction\}

Here are the outputs of the models:

[

\quad\{"model": "model\_1","answer": "\{output\_1\}"\},

\quad\{"model": "model\_2","answer": "\{output\_2\}"\}

]

Now please rank the models by the quality of their answers, so that the model with rank 1 has the best output. Then return a list of the model names and ranks, i.e., produce the following output:

[

\quad\{'model': <model-name>, 'rank': <model-rank>\},

\quad\{'model': <model-name>, 'rank': <model-rank>\}

]

Your response must be a valid Python dictionary and should contain nothing else because we will directly execute it in Python. Please provide the ranking that the majority of humans would give.

\end{tcolorbox}

The results of the win rate after v.s. before preference alignment are shown in Table~\ref{tab:llm eval}. We found that on all models and datasets, generators that have undergone RAG preference alignment have better response informativeness and response robustness.

\section{Case Study}
\label{appdix:case study}

To more intuitively demonstrate the improvements of preference optimization on response informativeness, response robustness, and citation quality, we randomly sampled several outputs from the \textsc{Llama2-13b-chat} before and after preference alignment (PA) for comparison. The examples are as follows.

Take Case 1 as an example, we found that the generator without PA-RAG failed to utilize the document containing the answer and was misled by the noisy document, mistakenly identifying Erika Flores as the actress who played Ingrid, resulting in the incorrect citation. In contrast, the generator with PA-RAG training correctly used the document containing the answer and provided a correct response, demonstrating better response informativeness, response robustness, and citation quality.

\begin{figure*}[t]
\begin{tcolorbox}[title=Case 1: Sampled from the ASQA]
\small
\textbf{Question:} Who played ingrid on dr quinn medicine woman?

\textbf{Documents:}

[1] she was featured throughout the sixth season, when her character fell in love with Jake Slicker. The role of Anthony (Grace and Robert E.'s adopted son) was played by Brenden Jefferson for four episodes in season four. He was replaced by Brandon Hammond, who continued in the role throughout seasons five and six. Jennifer Youngs did not begin playing Ingrid until the character's second appearance; the first time the character appeared, she was played by Ashley Jones. "Dr. Quinn" was best known for its large, supporting cast, and its high-concept storytelling. The series often used its semi-historical setting as a

[2] Erika Flores Erika Flores (born November 22, 1979 in Grass Valley, Nevada County, California) is an American former child actress. She is known for her role as the first Colleen Cooper in the "Dr. Quinn, Medicine Woman" TV series; Flores left the role in 1995, in the middle of the third season, and the role was then recast with Jessica Bowman taking over as Colleen. Flores guest-starred on the "" episode "". She played Marissa Flores, one of three young science-fair winners aboard the USS "Enterprise". In January 2009, Flores guest starred in "Big Baby", a fifth-season episode of the

[3] Heidi Kozak Heidi Karin Kozak (born June 22, 1963) is a Danish-born American actress. She is perhaps best known for her roles in several late-1980s horror films, including "Slumber Party Massacre 2", "" (both 1987), and "Society" (1989). She also starred in the first season of "Dr. Quinn Medicine Woman" in the supporting role of Emily Donovan. Although she received an on-screen departure, it is unknown why she was written out of the series. Additionally, she appears in the behind-the-scenes documentary "Sleepless Nights: Revisiting the Slumber Party Massacres". Starting in 2005 she began charity work that involved baking different kinds

[4] Jessica Bowman Jessica Robyn Bowman (born November 26, 1980) is an American actress known for her role as Colleen Cooper on "Dr. Quinn, Medicine Woman". Bowman was born in Walnut Creek, California. She began acting in community theater and commercials, and appeared as Darcy on "The Road Home". After guest appearances on "Boy Meets World", "NYPD Blue", and "Baywatch", Bowman assumed the role of Colleen Cooper on the popular drama "Dr Quinn, Medicine Woman" in season three, episode 15. She took over this role after the original portrayer, Erika Flores, did not renew her contract. Bowman won the Best Performance

\textbf{Short answers:} ['Ashley Jones',Jennifer Youngs']

\textbf{Output from the generator without PA:} Erika Flores played Ingrid on "Dr. Quinn, Medicine Woman"[2][3].

\textbf{Output from the generator with PA:} Ashley Jones and Jennifer Youngs both played the character Ingrid on "Dr. Quinn, Medicine Woman."[1]. Ashley Jones portrayed Ingrid in her first appearance, while Jennifer Youngs took over the role for subsequent appearances[1]. Additionally, Erika Flores played the first Colleen Cooper in the series before Jessica Bowman took over the role[2].
\end{tcolorbox}
\end{figure*}

\begin{figure*}[t]
\begin{tcolorbox}[title=Case 2: Sampled from the WebQuestions]
\small
\textbf{Question:} where did bristol palin go to school?

\textbf{Documents:}

[1] aunt and uncle where she attended West Anchorage High School. After returning to Wasilla, she attended and then graduated from Wasilla High School in May 2009. Palin first became the subject of media attention when her pregnancy was announced during her mother's unsuccessful run for Vice President. In February 2009, she told Fox News that abstinence is "not realistic at all," but that she would like it to become more accepted among people her age. At age 18 in May 2009, Palin appeared on the "Today" show and "Good Morning America" in recognition of the National Day to Prevent Teen

[2] season, where she was eliminated in the fourth week of competition. Palin was born on October 18, 1990, in Wasilla, Alaska to parents Todd and Sarah (née Heath) Palin. She was named "Bristol" for the Bristol Inn where her mother Sarah was employed; Bristol, Connecticut, the headquarters city of ESPN, where her mother Sarah hoped to work as a sportscaster; and the Bristol Bay area of Alaska, where her father Todd grew up. Palin was raised largely in Wasilla and attended Juneau-Douglas High School when her mother was governor of Alaska. During 2008, she briefly lived in Anchorage with her

[3] Wasilla High School Wasilla High School (WHS) is a public secondary school in Wasilla, Alaska, United States, serving students in grades 9–12. The school is part of the Matanuska-Susitna Borough School District, with admission based primarily on the locations of students' homes. The school earned widespread media attention in 2008 following former pupil Governor Sarah Palin's nomination as the Republican vice-presidential running mate to John McCain in the 2008 United States presidential election. As of 22 February 2018, there are fifteen AP classes offered. WHS also participates in the University of Alaska Anchorage's Tech Prep program, which allows students to

[4] school (Palmer Jr. Middle School) and one high school (Palmer High School). Palmer High is of a modest size with about 900 students and approximately 60 staff members. In 1999, Palmer High became the first school in Alaska to offer the International Baccalaureate program. About ten minutes from Palmer are "cross-town" rivals, Colony Middle and High Schools. Palmer is also home to Alaska Bible College, Alaska's only accredited school of theology and ministry, and Matanuska-Susitna College, an extended college of the University of Alaska Anchorage. There is also the Palmer Alaska Job Corps, 800 E. Lynn Martin Drive, across from

[5] Pacific University in Honolulu for a semester in the fall of 1982 and then to North Idaho College, a community college in Coeur d'Alene, for the spring and fall semesters of 1983. She enrolled at the University of Idaho in Moscow for an academic year starting in August 1984 and then attended Matanuska-Susitna College in Alaska in the fall of 1985. Palin returned to the University of Idaho in January 1986 and received her bachelor's degree in communications with an emphasis in journalism in May 1987. In June 2008, the Alumni Association of North Idaho College gave Palin its Distinguished

\textbf{Short answers:} ['Wasilla High School', 'West Anchorage High School', 'Juneau-Douglas High School']

\textbf{Output from the generator without PA:} Bristol Palin attended West Anchorage High School and graduated from Wasilla High School in May 2009[1][3].

\textbf{Output from the generator with PA:} Bristol Palin attended West Anchorage High School and later graduated from Wasilla High School in May 2009[1]. Additionally, she briefly lived in Anchorage and attended Juneau-Douglas High School during her mother's governorship of Alaska[2].
\end{tcolorbox}
\end{figure*}

\begin{figure*}[t]
\begin{tcolorbox}[title=Case 3: Sampled from the Natural Questions]
\small
\textbf{Question:} name the process of fusion of an egg with a sperm

\textbf{Documents:}

[1] the oocyte. Fusion between the oocyte plasma membranes and sperm follows and allows the sperm nucleus, the typical centriole, and atypical centriole that is attached to the flagellum, but not the mitochondria, to enter the oocyte. The protein CD9 likely mediates this fusion in mice (the binding homolog). The egg "activates" itself upon fusing with a single sperm cell and thereby changes its cell membrane to prevent fusion with other sperm. Zinc atoms are released during this activation. This process ultimately leads to the formation of a diploid cell called a zygote. The zygote divides to form a blastocyst and,

[2] fertilization process. Human fertilization Human fertilization is the union of a human egg and sperm, usually occurring in the ampulla of the fallopian tube. The result of this union is the production of a zygote cell, or fertilized egg, initiating prenatal development. Scientists discovered the dynamics of human fertilization in the nineteenth century. The process of fertilization involves a sperm fusing with an ovum. The most common sequence begins with ejaculation during copulation, follows with ovulation, and finishes with fertilization. Various exceptions to this sequence are possible, including artificial insemination, "in vitro" fertilization, external ejaculation without copulation, or copulation shortly

[3] and once bound together, the sperm and the egg fuse together to form two pronuclei These pronuclei supply the zygote with the genetic material necessary for the formation of an embryo. Additionally, once the fusion of the sperm and oocyte is complete, phospholipase C zeta is released from the sperm. As the sperm arrive at the egg, it is the egg that chooses the sperm and pulls it towards her. The selected sperm actually tries to swim away from the egg but is tethered to the egg by female hormones. The membrane around the egg literally opens up and swallows

[4] (microgametophyte) containing two haploid generative (sperm) cells and a tube nucleus. When a pollen grain makes contact with the female stigma, the pollen grain forms a pollen tube that grows down the style into the ovary. In the act of fertilization, a male sperm nucleus fuses with the female egg nucleus to form a diploid zygote that can then develop into an embryo within the newly forming seed. Upon germination of the seed, a new plant can grow and mature. The adaptive function of meiosis is currently a matter of debate. A key event during meiosis in a diploid cell

[5] Human fertilization Human fertilization is the union of a human egg and sperm, usually occurring in the ampulla of the fallopian tube. The result of this union is the production of a zygote cell, or fertilized egg, initiating prenatal development. Scientists discovered the dynamics of human fertilization in the nineteenth century. The process of fertilization involves a sperm fusing with an ovum. The most common sequence begins with ejaculation during copulation, follows with ovulation, and finishes with fertilization. Various exceptions to this sequence are possible, including artificial insemination, "in vitro" fertilization, external ejaculation without copulation, or copulation shortly after ovulation.

\textbf{Short answers:} [['fertilization'], ['Human fertilization'], ['fertilization'], ['fertilization']]

\textbf{Output from the generator without PA:} The process of fusion of an egg and a sperm involves several stages[1]. The fusion between the oocyte plasma membranes and sperm occurs, which allows the sperm nucleus, mitochondria, and centriole to enter the oocyte [1]. This process leads to the formation of a diploid cell called a zygote [1]. The zygote divides to form a blastocyst, marking the beginning of prenatal development [1]. The sperm and egg fuse together to form two pronuclei, supplying the zygote with the genetic material necessary for the formation of an embryo [3]. This fusion ultimately results in the formation of a fertilized egg, known as zygote, initiating prenatal development [2][5].

\textbf{Output from the generator with PA:} The process of fusion of an egg with a sperm is called fertilization[2]. This process involves the union of a human egg and sperm, typically occurring in the ampulla of the fallopian tube, resulting in the production of a zygote cell, or fertilized egg, which initiates prenatal development [2][5]. The fusion between the oocyte plasma membranes and sperm allows the sperm nucleus, centriole, and other cellular structures to enter the oocyte, ultimately leading to the formation of a diploid cell called a zygote [1]. This fusion is crucial for the transfer of genetic material from the sperm to the egg, enabling the formation of an embryo [3].
\end{tcolorbox}
\end{figure*}

\begin{figure*}[t]
\begin{tcolorbox}[title=Case 4: Sampled from the Natural Questions]
\small
\textbf{Question:} who was kat slater's sisters in eastenders

\textbf{Documents:}

[1] Slater family (EastEnders) The Slater family are a fictional family from the BBC soap opera "EastEnders", that have appeared since episode 1997, first broadcast in September 2000. The family are introduced as a central family unit consisting of father Charlie Slater (Derek Martin), grandmother Mo Harris (Laila Morse), Charlie's daughters, Kat Slater (Jessie Wallace), Lynne Slater (Elaine Lordan), Little Mo Morgan (Kacey Ainsworth) and Zoe Slater (Michelle Ryan), and Lynne's partner Garry Hobbs (Ricky Groves). A fifth sister, Belinda Peacock (Leanne Lakey), was introduced a year later. The family were the eighth to be introduced in the soap's history, replacing

[2] Kat Slater Kathleen "Kat" Moon (also Slater) is a fictional character from the BBC soap opera "EastEnders", played by Jessie Wallace. She was also played by Kate Peck in a flashback in 2001. Kat is the daughter of Viv Slater and Charlie Slater (Derek Martin), and arrived with her sisters Lynne Hobbs (Elaine Lordan) and Little Mo Morgan (Kacey Ainsworth). Her daughter Zoe Slater (Michelle Ryan) is initially believed to be her sister as well until Kat reveals she was raped by her uncle Harry Slater (Michael Elphick) as a child. She later marries Alfie Moon (Shane Richie). It is

[3] ends of the country, and in some cases beyond." Slater family (EastEnders) The Slater family are a fictional family from the BBC soap opera "EastEnders", that have appeared since episode 1997, first broadcast in September 2000. The family are introduced as a central family unit consisting of father Charlie Slater (Derek Martin), grandmother Mo Harris (Laila Morse), Charlie's daughters, Kat Slater (Jessie Wallace), Lynne Slater (Elaine Lordan), Little Mo Morgan (Kacey Ainsworth) and Zoe Slater (Michelle Ryan), and Lynne's partner Garry Hobbs (Ricky Groves). A fifth sister, Belinda Peacock (Leanne Lakey), was introduced a year later. The family were the

[4] Discussing the character's death, Digital Spy's Kris Green praised Crace's acting. He compared Danielle's revelation that Ronnie was her mother to the "EastEnders" storyline several years previously, when Kat Slater (Jessie Wallace) revealed she was actually the mother of her supposed sister Zoe (Michelle Ryan). Green opined: "Although it didn't quite capture the magic of [the Kat-Zoe scene] it definitely comes an "extremely" close second." He concluded: "I'd probably go as far as to say that it's one of the best episodes "EastEnders" has produced in a long time". The episode was selected as recommended viewing by "The Guardian" Sarah

[5] that she wants to get back together with him, but he tells her that he is not ready as he does not want to hurt her again. This causes Kat to reevaluate her life, so she visits the convent where she gave birth to Zoe to seek some answers, though it is now a nun's care home. A nun, Sister Ruth (June Whitfield), remembers her going into labour and they talk about the past and the future. Kat then leaves and goes to the police station to make a statement about Harry. Ruth checks their records and discovers that Kat

\textbf{Short answers:} [['Lynne', 'Little Mo'], ['Lynne', 'Little Mo', 'Zoe'], ['Lynne', 'Little Mo', 'Zoe']]

\textbf{Output from the generator without PA:} Kat Slater had several sisters in EastEnders[1][3]. She was introduced as part of the central family unit consisting of father Charlie Slater and his daughters, including Lynne Slater and Zoe Slater [1][2][3]. Additionally, a fifth sister, Belinda Peacock, was introduced a year after Kat's arrival [1].

\textbf{Output from the generator with PA:} Kat Slater had four sisters in "EastEnders."[1]. Her sisters were Lynne Slater, Little Mo Morgan, Zoe Slater, and later, Belinda Peacock[1]. Kat, portrayed by Jessie Wallace, was the daughter of Viv Slater and Charlie Slater, and she arrived with her sisters Lynne Hobbs and Little Mo Morgan[2]. Zoe Slater, played by Michelle Ryan, was initially believed to be her sister until Kat revealed that she was her daughter[2]. Belinda Peacock was introduced a year later[1].
\end{tcolorbox}
\end{figure*}

\begin{figure*}[t]
\begin{tcolorbox}[title=Case 5: Sampled from the TriviaQA]
\small
\textbf{Question:} England won the 1966 World Cup beating Germany 4-2. Hurst scored 3 of England's goals who scored the other?

\textbf{Documents:}

[1] the game went to extra time. In the 98th minute, Hurst found himself on the scoresheet again; his shot hit the crossbar, bounced down onto the goal line, and was awarded as a goal. Debate has long raged over whether the ball crossed the line, with the goal becoming part of World Cup history. England's final goal was scored by Hurst again, as a celebratory pitch invasion began. This made Geoff Hurst the only player ever to have scored three times in a single World Cup final. BBC commentator Kenneth Wolstenholme's description of the match's closing moments has gone down

[2] over France in their final group game. Roger Hunt scores both of England's goals. 23 July 1966: England beat Argentina 1–0 in the World Cup quarter-final thanks to a goal by Geoff Hurst. 26 July 1966: England reach the World Cup final by beating Portugal 2–1 in the semi-final. Bobby Charlton scores both of England's goals. 30 July 1966: England win the World Cup with a 4–2 win over West Germany in extra time. Geoff Hurst scores a hat-trick, with Martin Peters scoring the other goal. !style="width:15em"|Competition!!style="width:15em"|Winner!!style="width:15em"|Runner-up Liverpool, FA Cup winners the previous season and league champions in 1964, won

[3] described at the time as a 4–4–2. The match is remembered for England's only World Cup trophy, Geoff Hurst's hat-trick – the first, and to date, only one ever scored in a FIFA World Cup Final – and the controversial third goal awarded to England by referee Gottfried Dienst and linesman Tofiq Bahramov. Both teams were strong throughout the tournament. Each won two and drew one of their three matches in the group stages. England did not concede a goal until their semi-final against Portugal. One of the enduring images of the celebrations in Wembley immediately after the game was

[4] 1962 FIFA World Cup. Ramsey confidently predicted that at the next tournament, England would win the trophy, and they did just that. The 1966 World Cup saw England win the World Cup in a controversial 4–2 victory over West Germany. The three goals scored by Geoff Hurst within 120 minutes, of which some are controversial, are the only hat trick to be achieved in a World Cup final to date. Bobby Moore was the captain on that day, whilst Munich air crash survivor Bobby Charlton also played. Moore's West Ham colleagues Geoff Hurst and Martin Peters scored that day. The

[5] over Germany in the final held on home soil, at the old 1923 built Wembley stadium (since replaced by the new Wembley Stadium opened in 2007). In that final, Moore was the team captain and a central defender, while Wilson served as full back alongside him. Peters in midfield, and Hurst as one of the strikers, were the England goalscorers on the day of the final, with Hurst scoring the first goal in the first half, Peters scoring in the second half, and Hurst scoring two more in extra time to complete a famous World Cup hat-trick and secure victory.

\textbf{Short answers:} ['Martin Peters']

\textbf{Output from the generator without PA:} England won the 1966 World Cup by beating Germany 4-2[2]. Geoff Hurst scored 3 of England's goals [1].

\textbf{Output from the generator with PA:} Geoff Hurst scored 3 of England's goals in the 1966 World Cup final, while Martin Peters scored the other goal in England's 4-2 victory over Germany[2]. Hurst's hat-trick in the final remains the only one ever scored in a FIFA World Cup Final[3].
\end{tcolorbox}
\end{figure*}

\begin{figure*}[t]
\begin{tcolorbox}[title=Case 6: Sampled from the TriviaQA]
\small
\textbf{Question:} Who was the only British Prime Minister to hold office during the reign of King Edward VIII?

\textbf{Documents:}

[1] used government papers in breach of the "Fifty-year rule" then in operation. Cabinet Secretary Sir Norman Brook persuaded the Prime Minister to demand amendments to conceal the degree of Cabinet involvement in the abdication of King Edward VIII (who was still alive in 1961) and the degree to which the civil servants Horace Wilson and Warren Fisher had demanded that the former King "reorder his private life" afterwards. Former Prime Minister Lord Avon, who cherished his (somewhat exaggerated) reputation as an opponent of "appeasement", complained that such a book by a serving Cabinet Minister might be thought to express official

[2] not like: Wilhelm II. Edward's difficult relationship with his nephew exacerbated the tensions between Germany and Britain. In April 1908, during Edward's annual stay at Biarritz, he accepted the resignation of British Prime Minister Sir Henry Campbell-Bannerman. In a break with precedent, Edward asked Campbell-Bannerman's successor, H. H. Asquith, to travel to Biarritz to kiss hands. Asquith complied, but the press criticised the action of the King in appointing a prime minister on foreign soil instead of returning to Britain. In June 1908, Edward became the first reigning British monarch to visit the Russian Empire, despite refusing to visit in

[3] office of Prime Minister has coincided with the reigns of 11 British monarchs (including a Regency during the incapacity of George III from 1811 to his death in 1820), to whom the Prime Minister has been constitutionally head of government to the sovereign's headship of state. Until 1837 the death of a sovereign led to Parliament being dissolved within six months which led to a general election. Results of such elections were: Stanley Baldwin is the only Prime Minister to have served three sovereigns in successionKing George V, King Edward VIII and King George VI. Through being in office at

[4] Edward VIII Edward VIII (Edward Albert Christian George Andrew Patrick David; 23 June 1894 – 28 May 1972) was King of the United Kingdom and the Dominions of the British Empire, and Emperor of India, from 20 January 1936 until his abdication on 11 December the same year, after which he became the Duke of Windsor. Edward was the eldest son of King George V and Queen Mary. He was named Prince of Wales on his sixteenth birthday, nine weeks after his father succeeded as king. As a young man, he served in the British Army during the First World

[5] invested by King Edward VII at Buckingham Palace on 8 August 1902. (He was the only prime minister to be knighted during his term of office until Robert Menzies in 1963; various others were knighted after leaving the office; Sir Earle Page was already a knight when he briefly became prime minister in 1939.) He received an honorary LL.D. from the University of Cambridge in 1900, and honorary D.C.L. from the University of Oxford and LL.D. from the University of Edinburgh while visiting the United Kingdom in the summer of 1902. He also received the Freedom of the City of

\textbf{Short answers:} ['Stanley, 1st Earl Baldwin of Bewdley Baldwin', 'Stanley Baldwin, 1st Earl Baldwin of Bewdley, KG, PC', 'Stanley Baldwin', 'Stanley, 1st Earl Baldwin of Bewdley, Viscount Corvedale of Corvedale Baldwin', 'Power without responsibillity', 'Stanley Baldwin, 1st Earl Baldwin of Bewdley', 'Baldwin, Stanley, 1st Earl Baldwin of Bewdley']

\textbf{Output from the generator without PA:} Iain Macleod was the only British Prime Minister to hold office during the reign of King Edward VIII[1][3].

\textbf{Output from the generator with PA:} Stanley Baldwin was the only British Prime Minister to hold office during the reign of King Edward VIII[3]. Edward VIII reigned from 20 January 1936 until his abdication on 11 December the same year, after which he became the Duke of Windsor[4]. Stanley Baldwin served as Prime Minister during this period, making him the only Prime Minister to have served three sovereigns in succession: King George V, King Edward VIII, and King George VI[3].
\end{tcolorbox}
\end{figure*}

\begin{figure*}[t]
\begin{tcolorbox}[title=Case 7: Sampled from the TriviaQA]
\small
\textbf{Question:} Louis Smith won an individual silver medal in gymnastics at the 2012 London Olympics. For what discipline?

\textbf{Documents:}

[1] 2016 Rio Olympics, this time finishing behind teammate, and 2012 bronze medalist, Max Whitlock. Smith was also part of the Great Britain team that took the bronze in the men's artistic team all-around at the 2012 London Olympics. He is the only British gymnast to win Olympic medals in three separate Games, and only the second gymnast after Marius Urzică to win three successive Olympic pommel horse medals. A 2006 Commonwealth Games champion representing England in the pommel horse, in 2015 he became European champion, his first major international title representing Great Britain. Smith is a member of the Huntingdon

[2] Louis Smith (gymnast) Louis Antoine Smith, MBE (born 22 April 1989) is a retired British artistic gymnast. He received a bronze medal and two silver medals on the pommel horse at the 2008 Beijing Olympics, 2012 London Olympics and the 2016 Rio Olympics respectively, with the former marking the first time a British gymnast had placed in an Olympic event since 1928. He fell just short of gold in 2012, tying with Kristian Berki, but taking silver for a lower E or execution score. He followed this up with a second consecutive silver medal on the pommel horse at the

[3] but fell on his dismount, lowering his execution score. Despite this, he still placed 3rd behind Krisztián Berki and Cyril Tommasone to win the Bronze medal — the only medal won by the British men and women at those championships. In 2012, Smith won the silver medal in the pommel horse finals in the European Men's Artistic Gymnastics Championships. Smith was part of the Great Britain team at the London Olympics. The team won a bronze medal at the men's artistic team all-around event on 30 July 2012, marking the first team medal for a British Olympic gymnastics team in

[4] Horse at the World Cup in Moscow. On 9 August 2008, Smith qualified in fifth place for the Olympic final of the men's pommel horse, and on 17 August he won a bronze medal in the finals, becoming first British man to win a medal in the individual gymnastics at the Olympic Games since Walter Tysall won a silver medal in 1908 and the first Briton generally to win a medal in gymnastics at the Olympic Games since the women's team won a bronze medal in 1928. He is also the second black male gymnast to win a medal in

[5] was harder than Gymnastics as everything was a new challenge. I never expected to win – I just went out every week with the aim of doing my best." Smith subsequently participated in the 2014 Christmas special with guest professional, Aliona Vilani. The couple won after dancing a Quickstep that scored a perfect 10 from all judges. Louis Smith (gymnast) Louis Antoine Smith, MBE (born 22 April 1989) is a retired British artistic gymnast. He received a bronze medal and two silver medals on the pommel horse at the 2008 Beijing Olympics, 2012 London Olympics and the 2016 Rio Olympics

\textbf{Short answers:} ['Gymnastics pommel horse', 'Side Horse', 'Pommel Horse', 'Pommel horse', 'Pommel horse (gymnastics)', 'Pomel horse', 'Pommel-Horse']

\textbf{Output from the generator without PA:} Louis Smith won an individual silver medal in gymnastics at the 2012 London Olympics[5].

\textbf{Output from the generator with PA:} Louis Smith won an individual silver medal in gymnastics at the 2012 London Olympics for the pommel horse discipline[2]. He is a retired British artistic gymnast who received a bronze medal and two silver medals on the pommel horse at the 2008 Beijing Olympics, 2012 London Olympics, and the 2016 Rio Olympics, respectively[2].

\end{tcolorbox}
\end{figure*}

\begin{figure*}[t]
\begin{tcolorbox}[title=Case 8: Sampled from the TriviaQA]
\small
\textbf{Question:} Which Asian country hosted its first Formula One Grand Prix motor race in October 2010?

\textbf{Documents:}

[1] the inaugural teams in the ASEAN Basketball League which was founded in October 2009. Kranji Racecourse is run by the Singapore Turf Club and hosts several meetings per week, including international races—notably the Singapore Airlines International Cup. Singapore began hosting a round of the Formula One World Championship, the Singapore Grand Prix at the Marina Bay Street Circuit in 2008. It was the inaugural F1 night race, and the first F1 street race in Asia. It is considered a signature event on the F1 calendar. Kranji Racecourse is run by the Singapore Turf Club and hosts several meetings per week,

[2] 2011 Indian Grand Prix The 2011 Indian Grand Prix, formally the 2011 Formula 1 Airtel Grand Prix of India, was a Formula One motor race that was held on 30 October 2011 at the Buddh International Circuit in Greater Noida, Uttar Pradesh, India. It was the seventeenth round of the 2011 Formula One season and the first Formula One Grand Prix to take place in South Asia and first to take place in India. The 60-lap race was won by Red Bull Racing's Sebastian Vettel, after leading every lap of the race from pole position and setting the fastest lap

[3] 2010 Korean Grand Prix The 2010 Korean Grand Prix (formally the 2010 Formula 1 Korean Grand Prix) was a Formula One motor race held on 24 October 2010 at the Korea International Circuit in Yeongam, South Jeolla, South Korea. It was the seventeenth round of the 2010 Formula One season and the first Korean Grand Prix. The 55-lap race was won by Ferrari driver Fernando Alonso, who started from third position. Lewis Hamilton finished second for McLaren and Alonso's teammate Felipe Massa was third. This was the first time since the 2008 Australian Grand Prix that neither Red Bull was

[4] 2011 Korean Grand Prix The 2011 Korean Grand Prix, formally the 2011 Formula 1 Korean Grand Prix was a Formula One motor race that was held on 16 October 2011 at the Korea International Circuit in Yeongam, South Jeolla, South Korea. It was the sixteenth round of the 2011 Formula One season, the second running of the Korean Grand Prix, and the first race after Sebastian Vettel claimed the World Drivers' Championship. The 55-lap race was won by Vettel, after starting from second on the grid. Lewis Hamilton finished in second place for McLaren, and Mark Webber completed the podium

[5] held later than usual, to avoid a clash with the 2006 Commonwealth Games. For the first time, Bahrain hosted the first Grand Prix. Brazil hosted the last race, while Japan and China swapped their original dates. In 2006, the FIA announced the Belgian Grand Prix would not be part of the 2006 Formula One season, since the local authorities had started major repair work in Spa-Francorchamps. The Belgian Grand Prix returned in 2007, when Kimi Räikkönen took pole position and his 3rd Belgian Grand Prix win in a row. The calendar was initially announced as the same as for 2005,

\textbf{Short answers:} ['Nam Chosun', 'Korea, Republic Of', 'So. Korea', 'ISO 3166-1:KR', 'R.O.K.', 'Korea, Republic of', 'Korea, South', 'Daehan Minguk', 'Republic Of Korea', 'S. Korea', 'R. Korea', 'South Kroea', 'South korea', 'Republican Korea', 'Korea (R.O.K.)', 'The republic of korea', 'Korea South', 'ROK', 'Daehanminkook', 'South Korea/Economy', 'Daehanmingook', 'Korea (South)', 'Korea Republic', 'SKorea', 'S Korea', 'South-korea', 'Republican Corea', 'Republic of KOREA', 'Corea Republic', 'Daehanminkuk', 'Korea (Republic of)', 'Republic of korea', 'Korea (ROK)', 'South Koreans', 'Free Korea', 'Daehanminguk', 'South Corea', 'Modern Corea', 'South corea', "Taehan-min'guk", 'S.Korea', 'South-Korea', 'Republic of Corea', 'S. Koreans', 'S. Korean', 'South Korean', 'Korea (Seoul)', 'Namjoseon', 'Republic of Korea', 'Korea, S', 'Korean Republic', 'Republic of South Korea', 'South Korea', 'Corean Republic']

\textbf{Output from the generator without PA:} The Indian Grand Prix in October 2010 marked India's first Formula One Grand Prix motor race, making it the first Formula One Grand Prix to take place in South Asia and the first to take place in India [2].

\textbf{Output from the generator with PA:} South Korea hosted its first Formula One Grand Prix motor race in October 2010[3]. The 2010 Korean Grand Prix, formally the 2010 Formula 1 Korean Grand Prix, took place on 24 October 2010 at the Korea International Circuit in Yeongam, South Jeolla, South Korea, marking the first Korean Grand Prix in the country's history[3].
\end{tcolorbox}
\end{figure*}

\end{document}